\documentclass[sigconf,anonymous=false,9pt]{acmart}
\usepackage{graphicx}
\usepackage{subcaption}
\usepackage{multirow}
\usepackage{float}
\usepackage[belowskip=-6pt,aboveskip=3pt]{caption}
\usepackage{afterpage}
\usepackage{adjustbox}
\usepackage{xcolor}
\usepackage{tcolorbox}
\tcbuselibrary{listings,breakable}
\newtcolorbox{algorithmbox}[1][]{
  enhanced,
  breakable,
  colback=white,
  colframe=black,
  boxrule=0.8pt,
  sharp corners,
  fonttitle=\bfseries,
  title=Algorithm #1,
  listing only,
  listing options={basicstyle=\ttfamily\footnotesize, numbers=left, numberstyle=\tiny, stepnumber=1, numbersep=5pt}
}
\usepackage{mdframed}
\usepackage{longtable}
\usepackage{listings}
\usepackage{caption}
\AtBeginDocument{%
  }

\renewcommand\footnotetextcopyrightpermission[1]{}
\usepackage{titlesec}
\titlespacing*{\section}{0pt}{6pt}{3pt}
\titlespacing*{\subsection}{0pt}{4pt}{2pt}
\titlespacing*{\subsubsection}{0pt}{3pt}{1pt}
\titlespacing*{\paragraph}{0pt}{2pt}{1pt}

\usepackage{enumitem}
\setlist{nosep,topsep=0pt,partopsep=0pt,itemsep=0pt,parsep=0pt}

\usepackage{parskip}
\renewcommand\footnotetextcopyrightpermission[1]{} 
\begin{document}
\fancyfoot{}
\title{\textsc{GRID}: Graph-based Reasoning for Intervention and Discovery in Built Environments}

\author{Taqiya Ehsan}
\email{taqiya.ehsan@rutgers.edu}
\orcid{0009-0007-1300-5379}
\affiliation{%
 \institution{Rutgers University}
 \city{New Brunswick}
 \state{New Jersey}
 \country{USA}
}

\author{Shuren Xia}
\email{shuren.xia@rutgers.edu}
\orcid{0009-0007-7459-7080}
\affiliation{%
 \institution{Rutgers University}
 \city{New Brunswick}
 \state{New Jersey}
 \country{USA}
}

\author{Jorge Ortiz}
\email{jorge.ortiz@rutgers.edu}
\orcid{0000-0003-3325-1298}
\affiliation{%
 \institution{Rutgers University}
 \city{New Brunswick}
 \state{New Jersey}
 \country{USA}
}

\renewcommand{\shortauthors}{Ehsan et al.}

\begin{abstract}
Manual HVAC fault diagnosis in commercial buildings takes 8-12 h per incident and 
achieves 60\% diagnostic accuracy—a symptom of analytics that stop at correlation 
instead of causation. To close this gap, we present \textsc{GRID} 
(Graph-based Reasoning for Intervention and Discovery), a three-stage causal 
discovery pipeline that combines constraint-based search, neural structural equation 
modeling, and language model priors to recover directed acyclic graphs from building 
sensor data. Across six benchmarks—synthetic rooms, EnergyPlus simulation, ASHRAE Great 
Energy Predictor III dataset, and a live office testbed—\textsc{GRID} achieves F1 scores 
ranging from 0.65 to 1.00, with exact recovery (F1 = 1.00) in three controlled 
environments (Base, Hidden, Physical) and strong performance on real-world data 
(F1 = 0.89 ASHRAE, 0.86 noisy conditions). The method outperforms ten baseline approaches 
across all evaluation scenarios. Our intervention scheduling achieves low operational 
impact in most scenarios (cost $\leq$ 0.026) while reducing risk metrics compared to 
baseline approaches. The framework integrates constraint-based methods, neural architectures, 
and domain-specific language model prompts to address the observational-causal gap in 
building analytics.
\end{abstract}




\maketitle

\begin{figure*}[!t]
\centering
\begin{adjustbox}{width=\textwidth,center}
\begin{tabular}{ccccccc}
\begin{subfigure}[b]{0.16\textwidth}
\centering
\includegraphics[width=\textwidth]{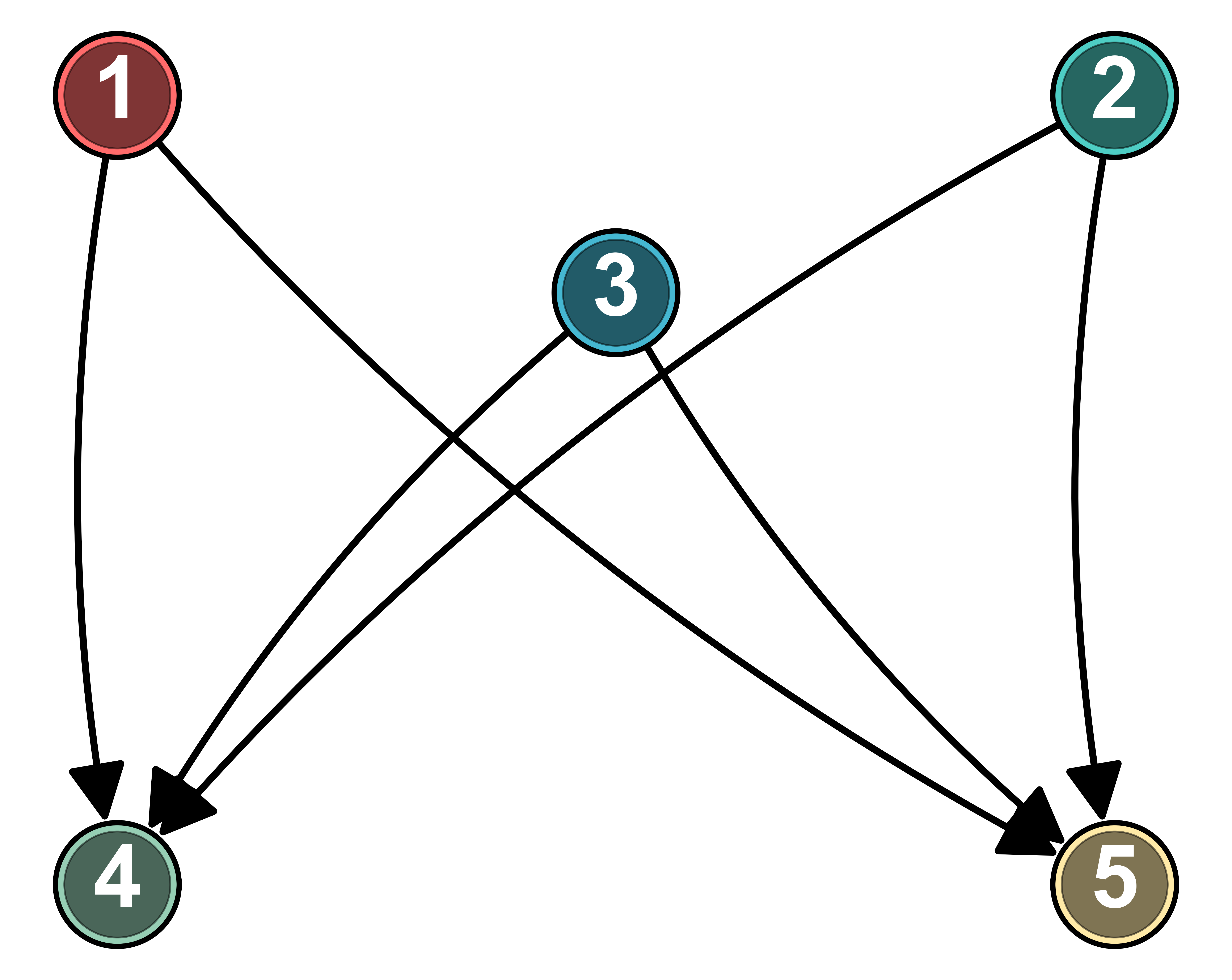}
\caption{\tiny Ground Truth}
\end{subfigure}
&
\begin{subfigure}[b]{0.16\textwidth}
\centering
\includegraphics[width=\textwidth]{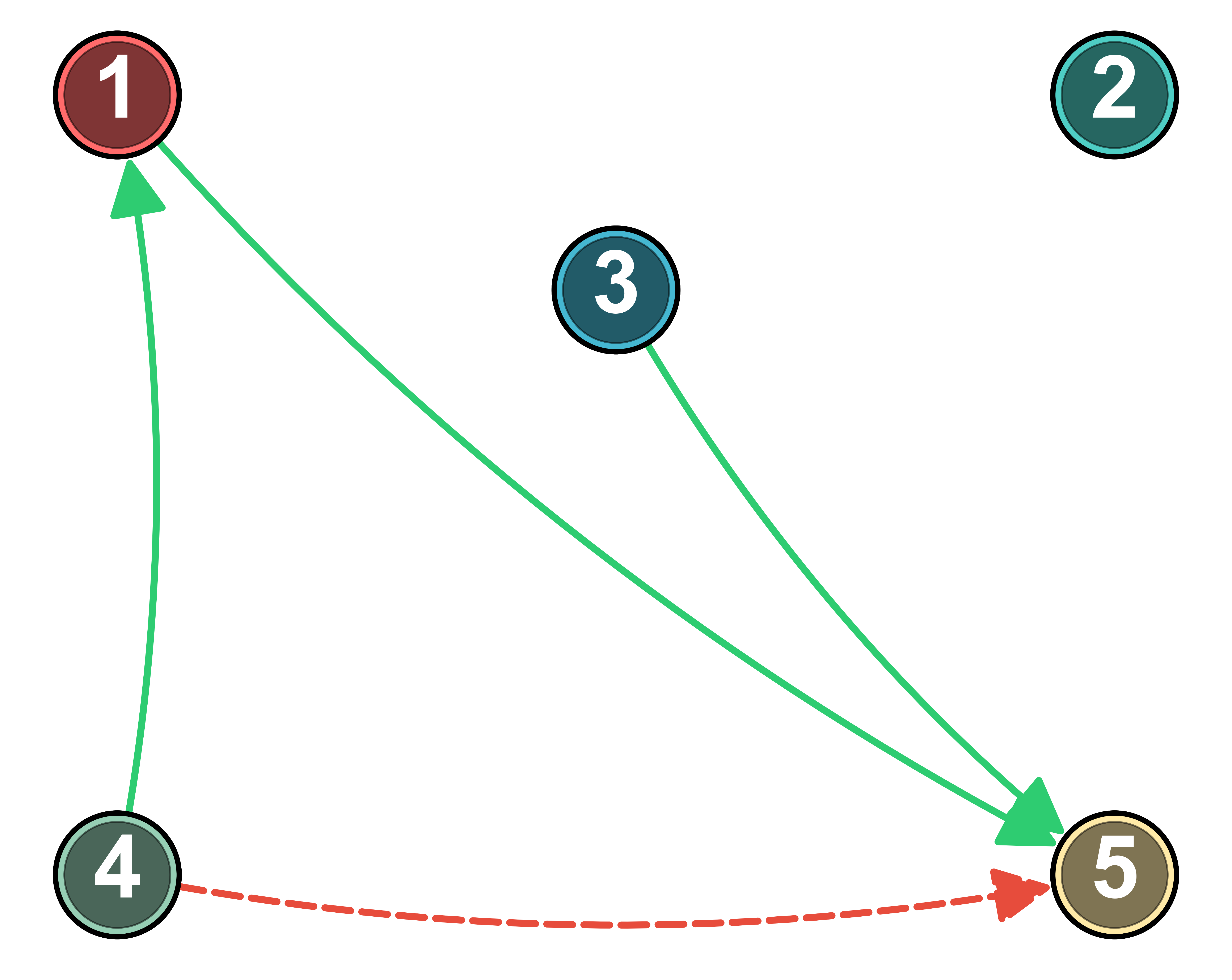}
\caption{\tiny GIES}
\end{subfigure}
&
\begin{subfigure}[b]{0.16\textwidth}
\centering
\includegraphics[width=\textwidth]{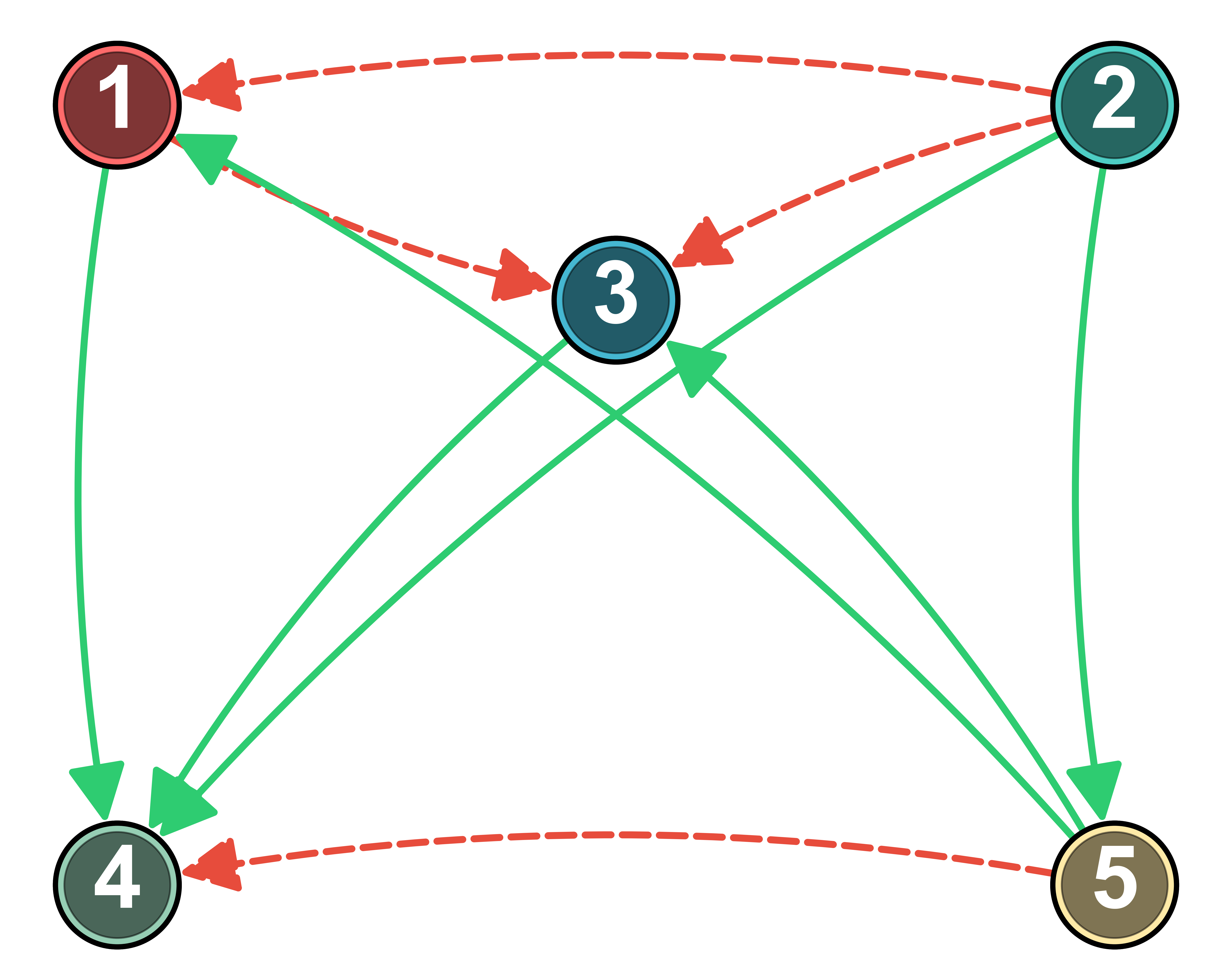}
\caption{\tiny JCI}
\end{subfigure}
&
\begin{subfigure}[b]{0.16\textwidth}
\centering
\includegraphics[width=\textwidth]{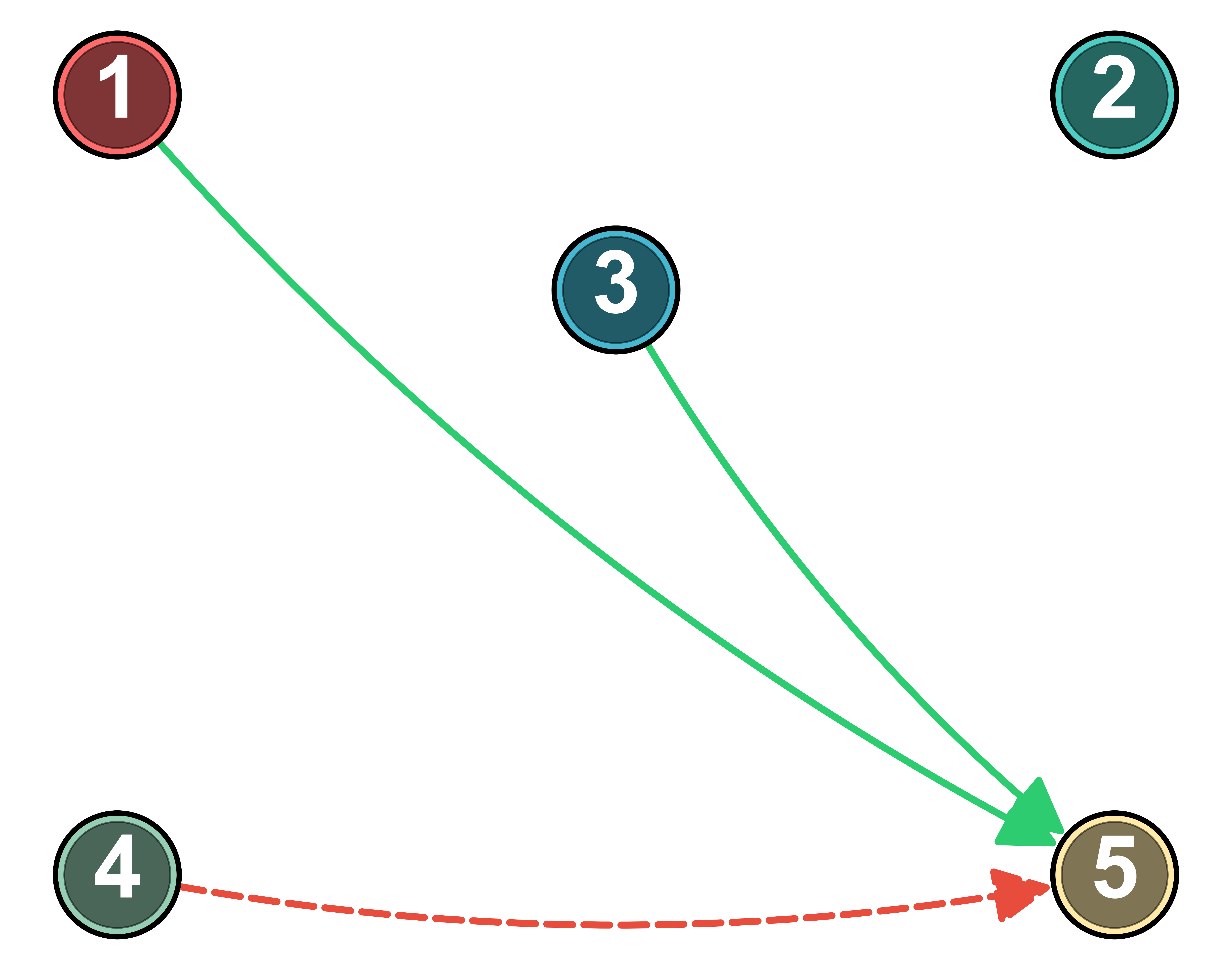}
\caption{\tiny ABCD}
\end{subfigure}
&
\begin{subfigure}[b]{0.16\textwidth}
\centering
\includegraphics[width=\textwidth]{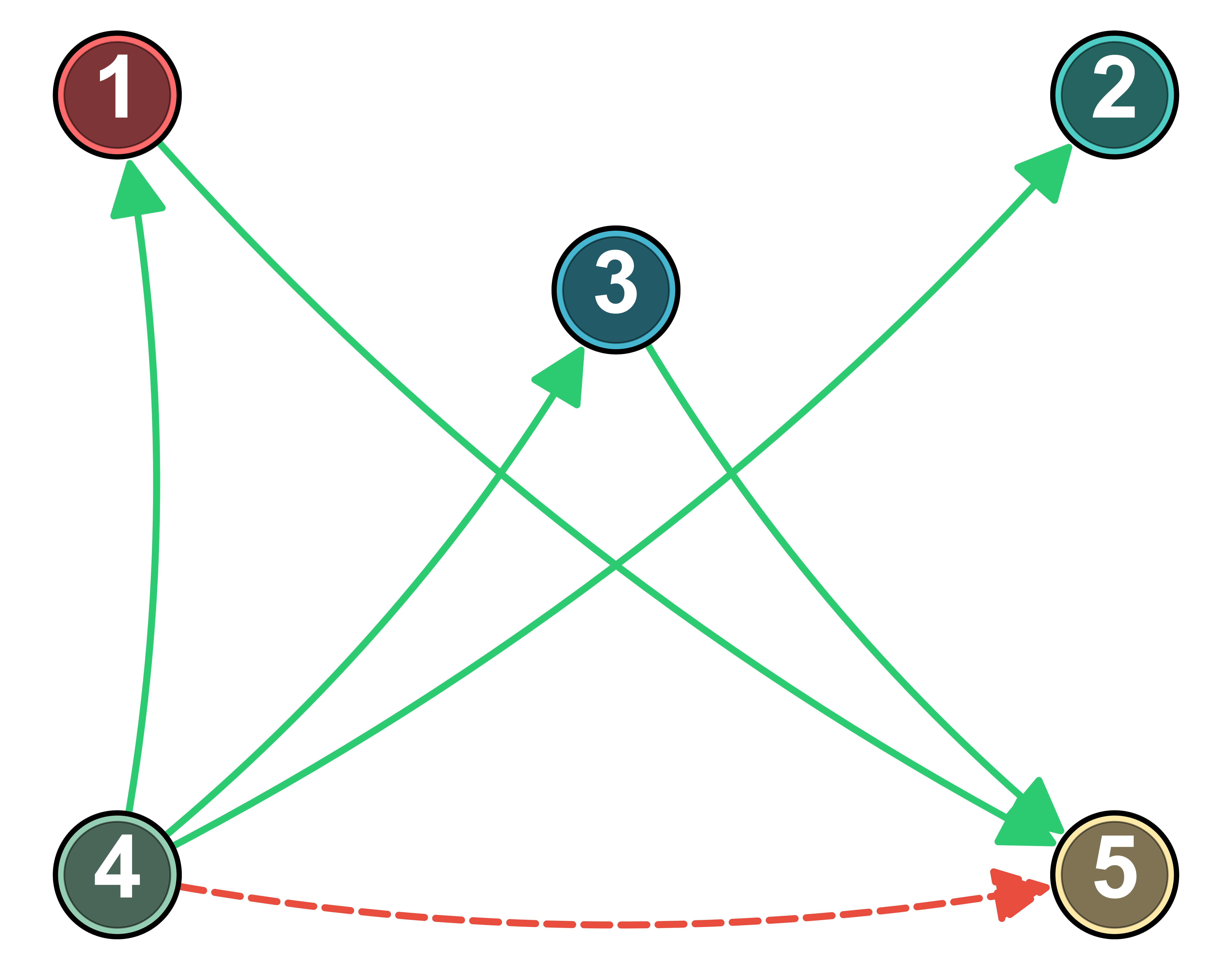}
\caption{\tiny Causal Bandits}
\end{subfigure}
&
\begin{subfigure}[b]{0.16\textwidth}
\centering
\includegraphics[width=\textwidth]{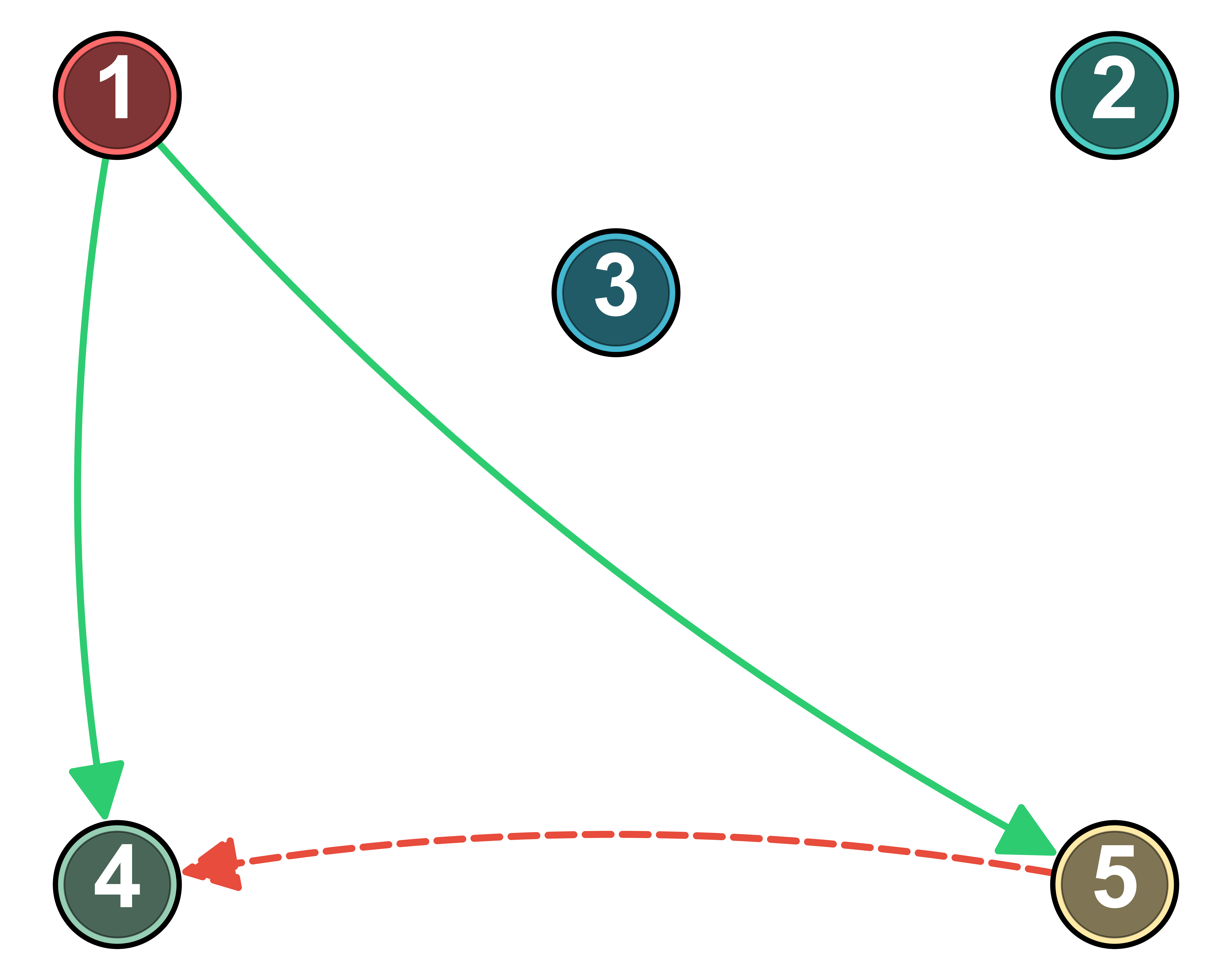}
\caption{\tiny ICP}
\end{subfigure} \\
\\[0.2cm] 
\begin{subfigure}[b]{0.16\textwidth}
\centering
\includegraphics[width=\textwidth]{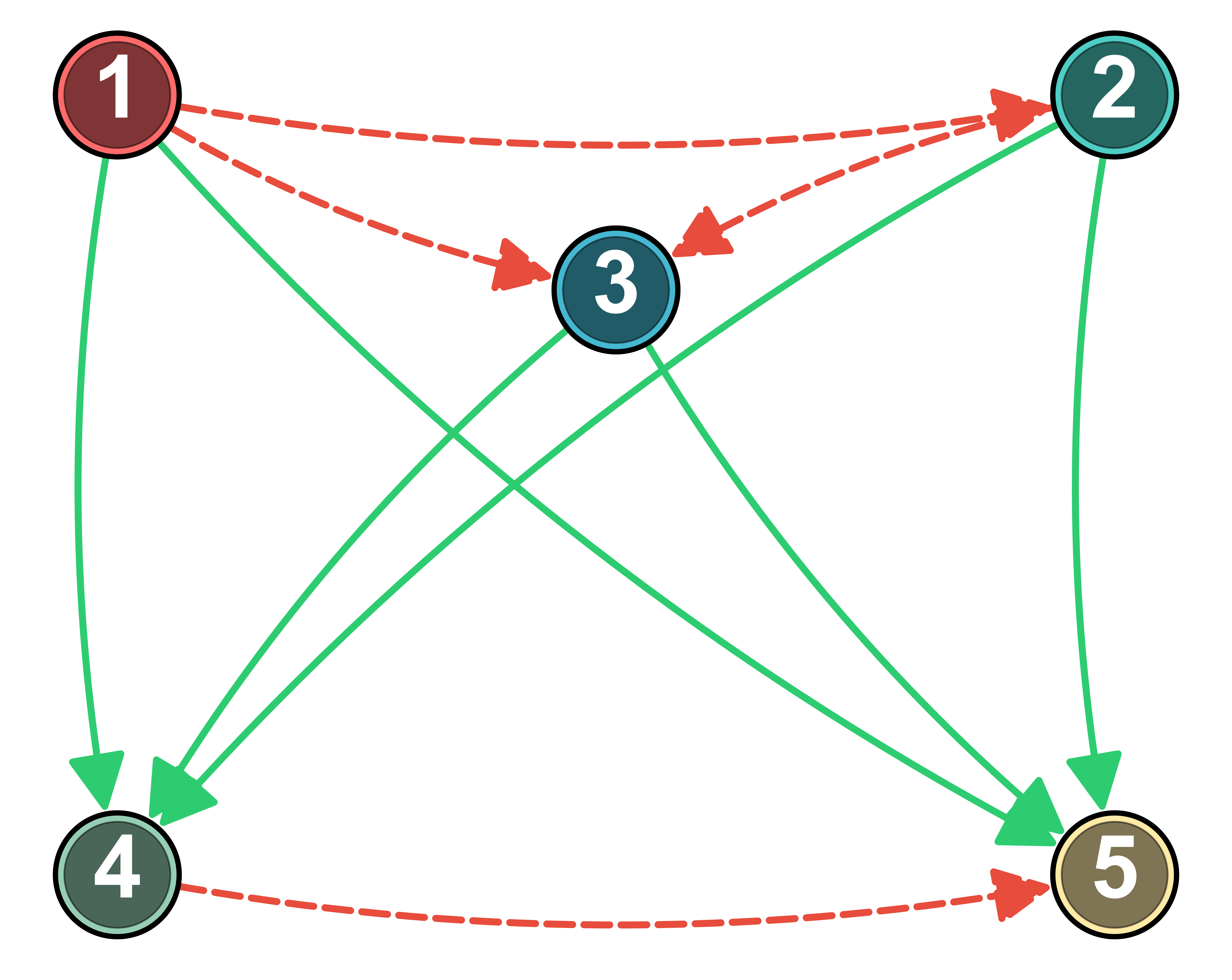}
\caption{\tiny IID}
\end{subfigure}
&
\begin{subfigure}[b]{0.16\textwidth}
\centering
\includegraphics[width=\textwidth]{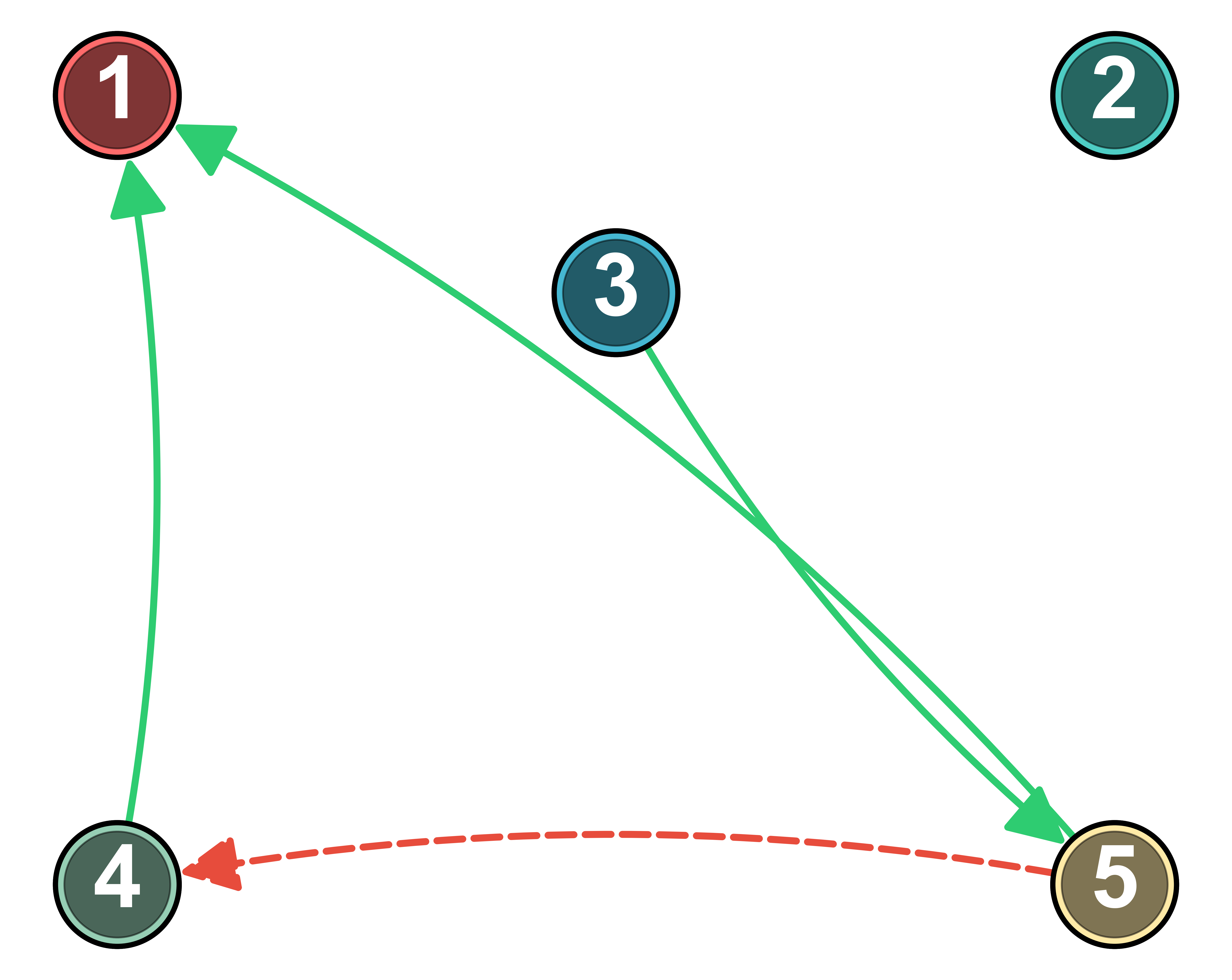}
\caption{\tiny NOTEARS}
\end{subfigure}
&
\begin{subfigure}[b]{0.16\textwidth}
\centering
\includegraphics[width=\textwidth]{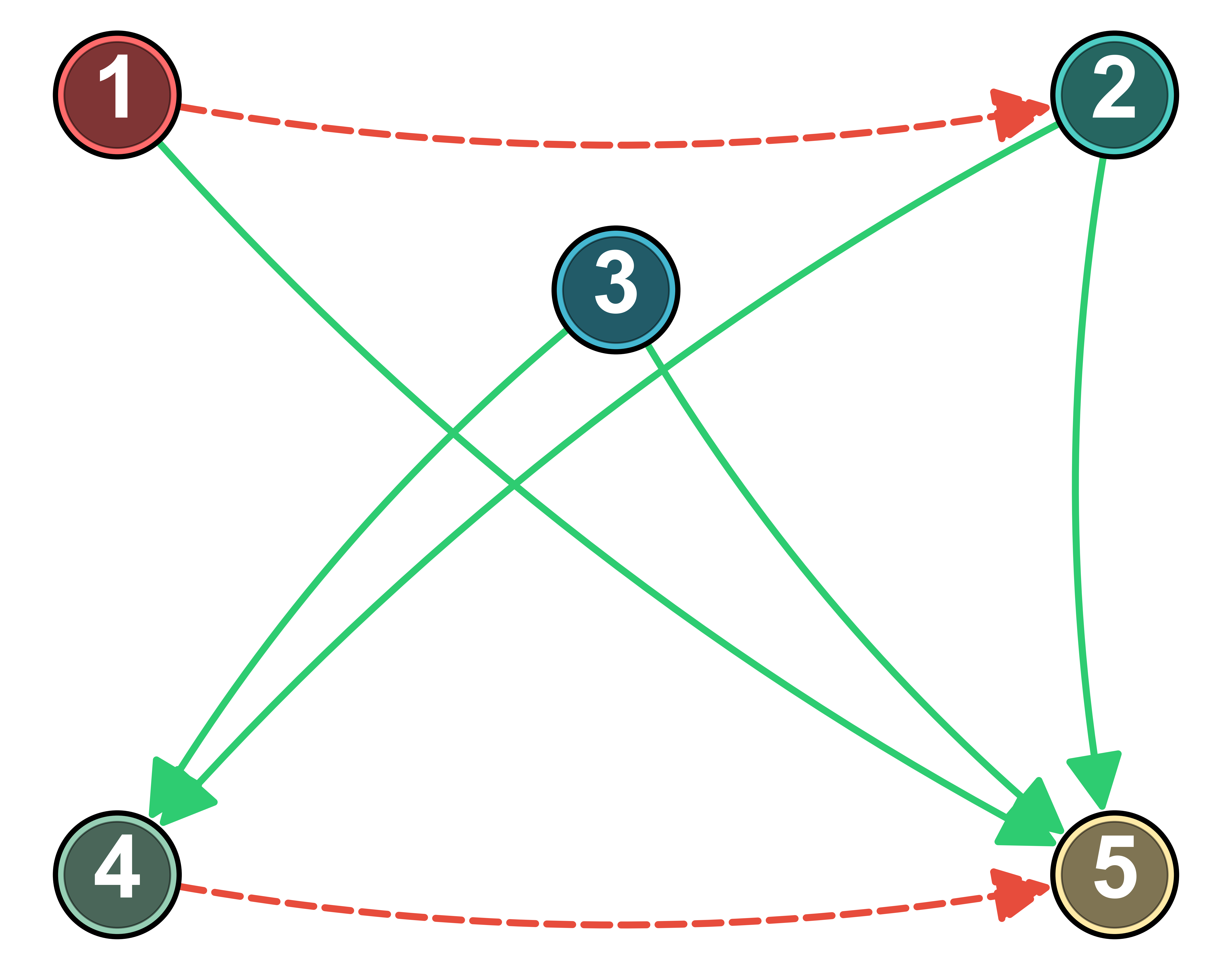}
\caption{\tiny PC}
\end{subfigure}
&
\begin{subfigure}[b]{0.16\textwidth}
\centering
\includegraphics[width=\textwidth]{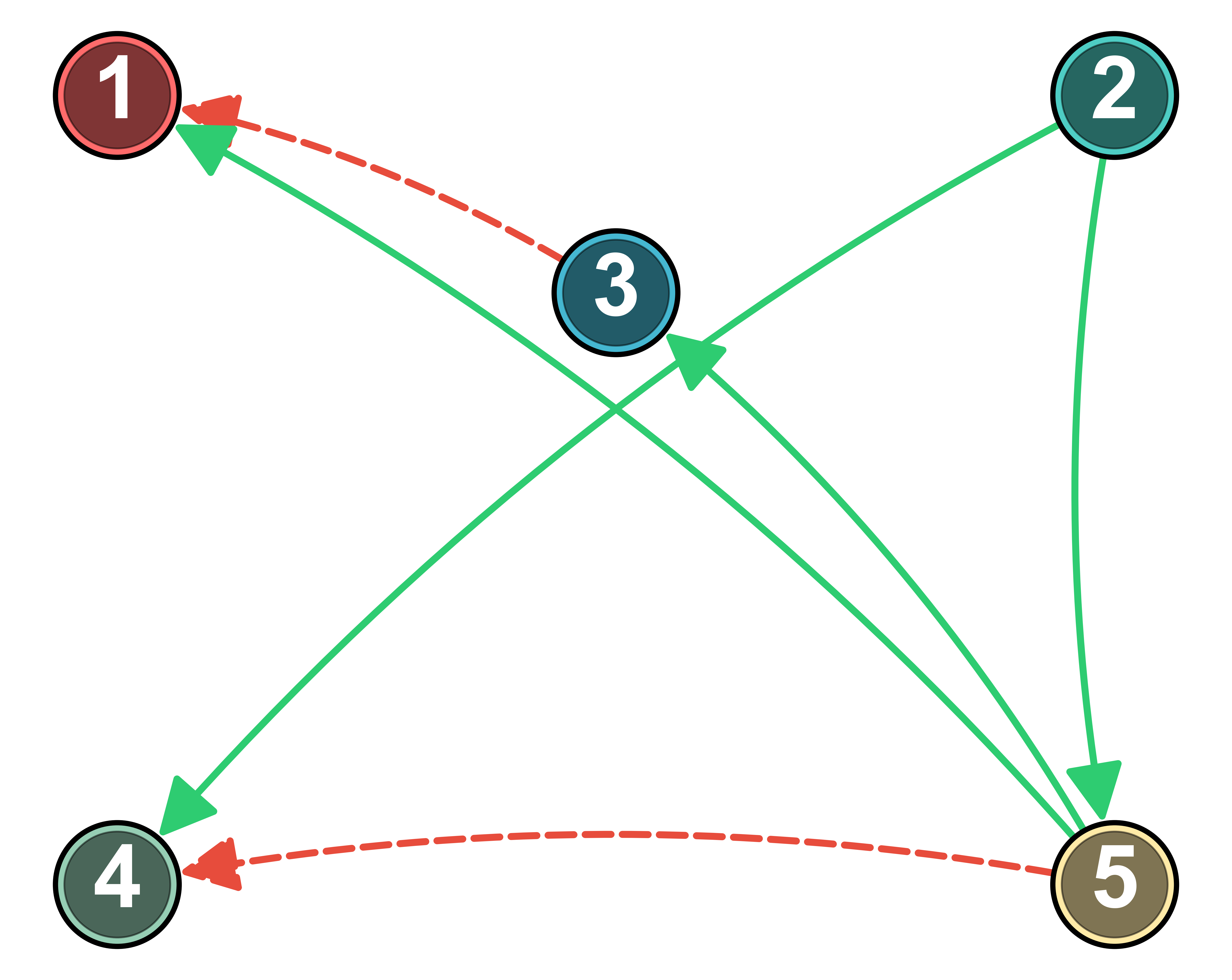}
\caption{\tiny SAM}
\end{subfigure}
&
\begin{subfigure}[b]{0.16\textwidth}
\centering
\includegraphics[width=\textwidth]{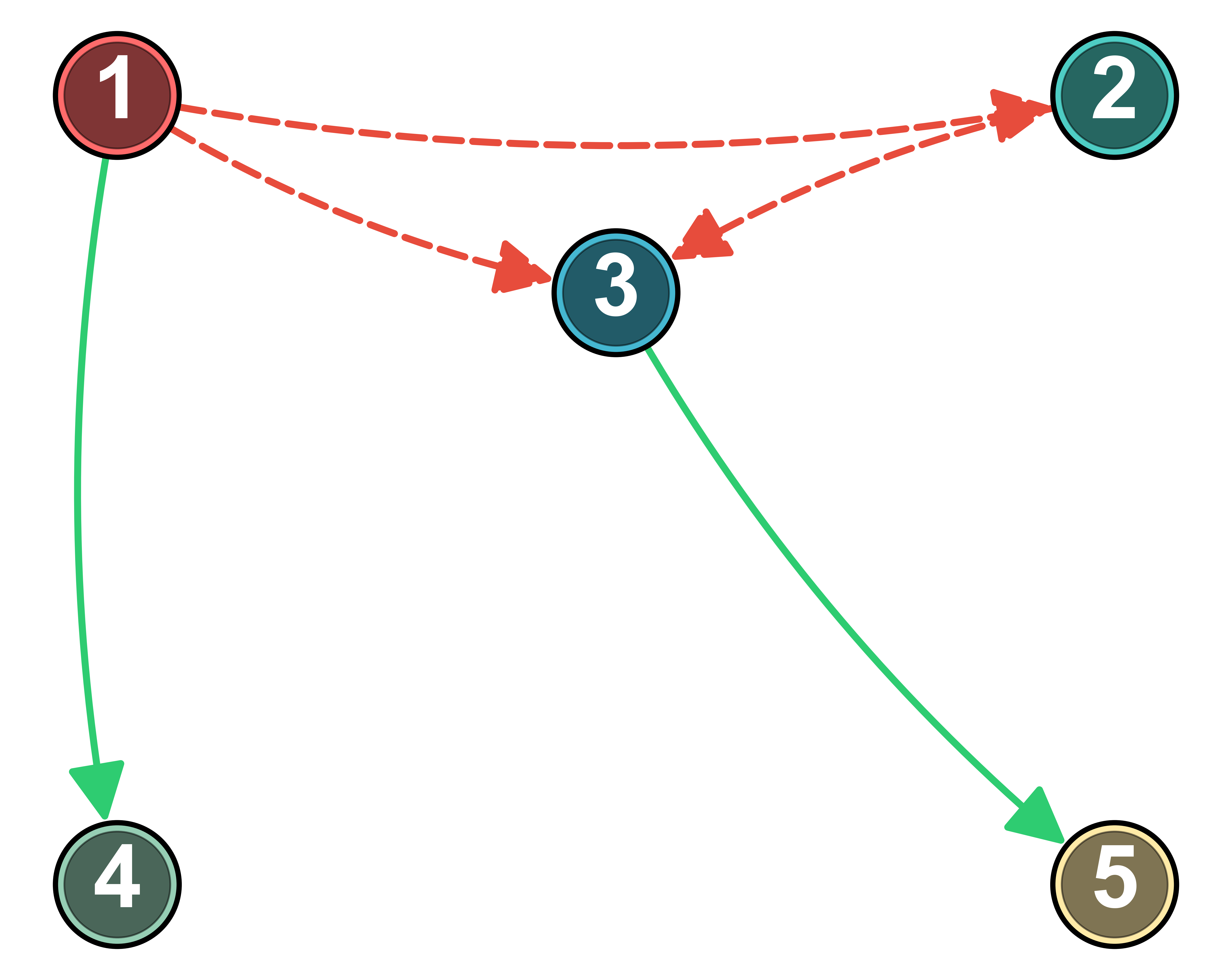}
\caption{\tiny LLM}
\end{subfigure}
&
\begin{subfigure}[b]{0.16\textwidth}
\centering
\includegraphics[width=\textwidth]{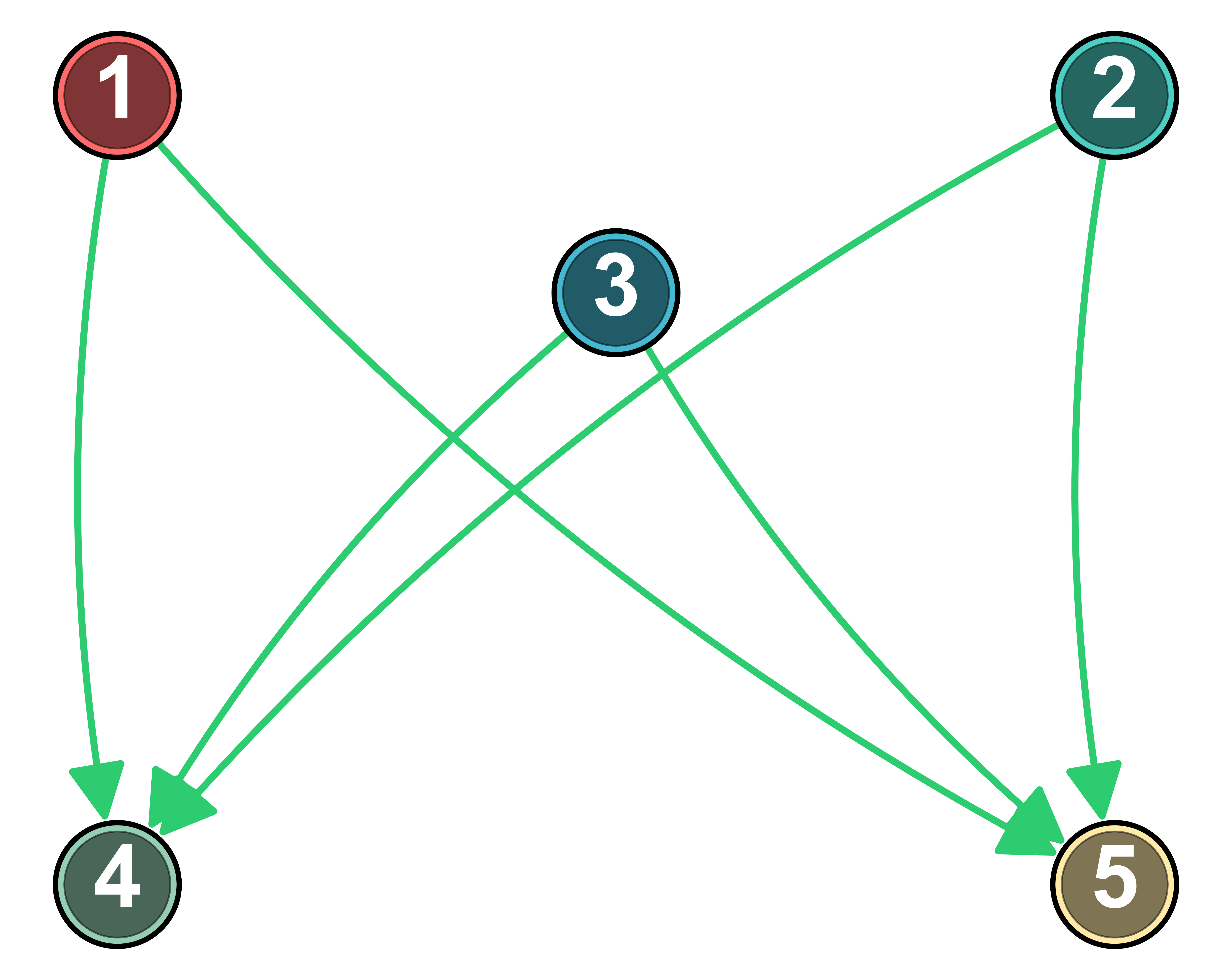}
\caption{\tiny \textbf{\textsc{GRID}}}
\end{subfigure}
&
\quad
\end{tabular}
\end{adjustbox}
\vspace{0.3cm}
\caption{
\textbf{Discovered causal DAGs by different methods in the \textit{Base Simulation} setup. The ground truth DAG (a) illustrates the true causal structure, while the other panels show outputs from benchmark methods in Section~\ref{benchmark_summary} and \textsc{GRID}. \textbf{Nodes are indexed as follows:} \textbf{1. temperature, 2. humidity, 3. air quality, 4. energy consumption, 5. overall satisfaction}. \textbf{Green solid edges indicate correct (true positive) causal relationships; red dashed edges indicate incorrect (false positive) ones}. While individual node connections may be visually dense, the key takeaway lies in the structural fidelity of each method. \textsc{GRID} achieves perfect reconstruction, matching the ground truth without spurious edges}.
}
\label{fig:base-dags}
\end{figure*}

\section{Introduction}

Buildings account for 30\% of global electricity consumption across all loads~\cite{iea2019global}, yet within this sector, HVAC faults can persist for days because most analytics pipelines rely on correlation, not causation. Field technicians spend 8-12 h isolating temperature anomalies and achieve 60\% diagnostic accuracy even in sensor-rich environments~\cite{kathirgamanathan2021data}. Missed faults contribute to 15\% of energy waste in buildings, costing the US \$50 billion annually~\cite{ogunkan2025exploring}.

\textsc{GRID} addresses this challenge through modular causal discovery that (i) combines constraint-based, neural, and language-guided methods under acyclicity constraints and (ii) validates uncertain edge relationships through targeted interventions. Our approach integrates observational learning with physical system knowledge to recover actionable causal structures from building sensor data.

Building environments present distinct challenges for causal discovery algorithms. Variables exhibit heavy confounding due to shared environmental drivers~\cite{5523244}. Physical constraints limit available interventions to accessible actuators. Data loss rates exceed 15\% in legacy deployments~\cite{czekster2022incorporating}. Existing building studies deploy these algorithms in isolation; none combine them with language-guided refinement or assesses intervention risk.

Through controlled experiments and real-world validation, \textsc{GRID} demonstrates performance across multiple environments. We evaluated on synthetic data, EnergyPlus~\cite{crawley2001energyplus} simulations, a public building dataset, and a physical office testbed. The framework outperforms baseline methods on structural accuracy metrics and intervention efficiency measures.


\textbf{Contributions.} GRID 1) merges constraint-based, neural, and LLM-guided causal discovery under acyclicity constraints, 
2) ranks edges to design low-disruption validation experiments, 
3) outperforms ten baselines across synthetic, simulation, real-world, and testbed data, and 
4) provides open-source access to code, datasets, and domain prompts. \textbf{GRID} is the first framework to integrate all three strands—constraint-based search, neural structure learning, and LLM-guided edge refinement—into a single loop that outputs a fully directed acyclic graph (DAG), supports intervention planning, and quantifies operational risk/cost. 

Building sensor networks generate measurements across variables such as temperature, humidity, air 
quality, energy consumption, and occupancy satisfaction. Physical constraints 
restrict interventions to available actuated systems per building. Variable dependencies 
follow non-linear patterns that constraint-based methods miss, while neural approaches may overfit to limited training data.

\textsc{GRID} employs three-stage causal discovery to address these limitations. 
The constraint-based stage identifies skeleton structures from observational data 
using the PC algorithm. The neural stage learns continuous dependencies between variables through structural equation modeling. The language model stage incorporates domain 
knowledge to refine edge orientations using building-specific prompts. Cross-method 
consensus guides edge validation and intervention prioritization.

\textbf{Novelty.} Prior building-analytics studies apply either constraint-based discovery alone~\cite{zhang2022fdd,chen2022design}, neural graph learners such as SAM without structural guarantees~\cite{kalainathan2018structural}, or couple statistical discovery with language models limited to post hoc explanation~\cite{long2023causal,neogi2025insightbuild}. \textsc{GRID} is the first to synergize constraint-based algorithms, neural causal models, and LLM-guided edge updates into a unified iterative process that outputs a complete DAG, plans interventions, and assesses operational cost and risk. No previous building-energy work combines these capabilities; comparable hybrids in the AI literature are recent and domain-agnostic~\cite{khatibi2024alcm,sunleveraging}.

The intervention scheduler reduces actuation requirements while maintaining discovery 
accuracy. We rank edges by confidence scores derived from cross-method agreement and 
target low-confidence relationships for validation. Each intervention tests specific 
causal hypotheses through systematic perturbations. The scheduler adapts to different actuator capabilities, from binary controls to continuous adjustments.

Our evaluation spans simulation logs and a week-long physical testbed, 
benchmarking GRID against ten constraint, neural, optimization, and 
LLM-based baselines. Metrics—F1, SHD, and normalized intervention cost—show 
consistent gains in both controlled and field conditions. 

Section \ref{problem_formulation} defines the task; Sections \ref{arch}–\ref{results} cover architecture, intervention scheduling, experimental setup, and results. The framework ingests time-series from our custom simulator and testbed sensors. Hybrid discovery plus targeted interventions yield actionable graphs for building operations. Code, datasets, and prompts will be released under an open-source license.

\section{Related Works} \label{related_works}
Recent years have seen increased interest in combining causal discovery with data-driven modeling for control and decision-making. In particular, large language models (LLMs), structural equation models (SEMs), and simulation-based approaches have been explored for reasoning about causal relationships in both abstract and physical domains~\cite{yu2024causaleval,neogi2025insightbuild}. This section reviews relevant research across five areas: causal discovery algorithms, validation strategies, evaluation metrics, applications in built environments, and HVAC systems.

\subsection{Causal Discovery Methods}



Causal discovery extracts causal graphs from data. PC~\cite{spirtes2000causation} uses conditional‐independence tests but degrades in noise or with hidden confounders~\cite{colombo2012learning,glymour2019review}. SEM methods such as SAM~\cite{kalainathan2018structural} model non-linear relations~\cite{rosseel2022structural,monti2020causal} yet demand precise specification and do not scale well. Deep-learning and LLM approaches mine structure from text~\cite{sunleveraging,long2023causal,zevcevic2023causal} but face interpretability and robustness limits in noisy domains. \textsc{GRID} integrates PC, SAM, and LLMs, combining symbolic, statistical, and model-based reasoning to withstand noise, accommodate latent variables, and validate discovered structures.

\subsection{Intervention-Based Validation}
Validating causal structure is essential in safety-critical settings like buildings. Pearl’s framework~\cite{pearl2009causality} emphasizes the need for interventional testing to distinguish causal from associational claims. Yet, real-world interventions can be costly or infeasible. While simulation-based validation and probing exist~\cite{meinshausen2016methods}, they are often decoupled from discovery.

\textsc{GRID} integrates an iterative validation loop, refining candidate graphs via targeted, low-cost simulated interventions. This process progressively confirms or removes uncertain edges, producing empirically grounded, robust structures. Additionally, \textsc{GRID} provides edge-level confidence scores to quantify causal certainty and support control decisions.

\subsection{Causal Graph Metrics and Evaluation}
Standard causal graph metrics include Structural Hamming Distance (SHD), precision, recall, and F1 score~\cite{zevcevic2023causal, pearl2009causality}. While SHD captures edge-level similarity, it overlooks the practical impact of incorrect edges in control tasks. Most benchmarks also assume full observability and ignore causal uncertainty.

\textsc{GRID} extends evaluation with edge confidence and intervention success rate, enabling more task-relevant and empirically grounded assessment. This supports not only structural comparison but also evaluation of behavioral accuracy under control.

\subsection{Applications in Building Systems}
Causal inference has increasingly informed building intelligence tasks like energy management, occupancy modeling, fault detection, and indoor environmental quality~\cite{chen2022introducing, zapata2024combining}, addressing limitations of purely correlational models in dynamic, partially observable settings.

Many existing methods rely on fixed or expert-defined causal structures and often omit uncertainty and empirical validation. While simulators like EnergyPlus~\cite{crawley2001energyplus} and TRNSYS~\cite{klein2000trnsys} offer high-fidelity forward models, they lack structure learning and intervention feedback, leaving causal claims unverified.

InsightBuild~\cite{neogi2025insightbuild} partly addresses this by combining Granger causality with LLM-generated explanations for energy anomalies, but without iterative interventions or formal validation.

\textsc{GRID} integrates dynamic discovery, intervention-based validation, and simulation feedback in a unified loop, enabling adaptive, statistically robust, and operationally relevant causal models across diverse building applications.

\subsection{Causal Reasoning in HVAC Systems}
HVAC systems are among the most complex and energy-intensive building subsystems, featuring non-linear dynamics, delays, and strong coupling with environmental and occupancy factors. Standard tests like ASHRAE 40~\cite{ashrae40} evaluate components in isolation, missing compounded effects from humidity, window states, and occupant behavior~\cite{de2011closing, burman2014review}.

Causal reasoning is crucial to identify true drivers of thermal outcomes; for instance, temperature rises may stem from solar load, occupancy, or HVAC faults, where uninformed interventions risk inefficiency. Modeling the causal graph enables tracing observed effects to their sources for targeted control.

\textsc{GRID} uses HVAC control as a canonical domain, combining latent variables, physical constraints, and actuation feedback. It validates edges via simulated interventions (e.g., airflow or setpoint changes), facilitating fine-grained fault diagnosis and robustness under uncertainty.

While prior work integrates causal reasoning with control, building-specific assumptions demand tailored methods. The next section reviews causal modeling principles to frame intervention-driven challenges in built environments.
\section{Problem Formulation} \label{problem_formulation}
To operationalize causal modeling within building automation pipelines, we formalize interactions between environmental, performance, and occupant-centric variables under both observational and interventional settings. Built environments are treated as dynamic cyber-physical systems in which physical processes (e.g., thermodynamics, ventilation) co-evolve with occupant behavior~\cite{yang2014systematic}. Although Building Management Systems (BMS) have become increasingly common, only 30\% of U.S. commercial floor spaces incorporate such systems as of 2018~\cite{eia2018cbecs}, leaving the majority of buildings either unmanaged or partially instrumented. We address this heterogeneity with \textsc{GRID} by generalizing across both sensor-rich and resource-constrained deployments, enabling causal inference from variable-quality data and supporting modular interventions.

\subsection{Modeling Objective}
Through \textsc{GRID}, we aim to recover the true causal graph $\mathcal{G}_{\text{true}} = (V, E_{\text{true}})$ over building variables $V$ using both observational data $\mathcal{D}_{\text{obs}} = \{X^{(i)}\}_{i=1}^N$ and direct interventions $\text{do}(X_j = x_j')$. The output is a DAG aligned with physical and temporal constraints. We assume causal sufficiency and full observability unless stated otherwise. We focus on variables central to building operation and occupant experience. Environmental inputs: \textit{temperature}, \textit{humidity}, and \textit{air quality} are standard in sensing systems~\cite{zhang2020systematic,saini2020comprehensive,ashrae-energy-prediction} and affect energy use and comfort. Outcomes include \textit{HVAC load}, \textit{energy consumption}, and \textit{occupant satisfaction}, estimated via ISO 7730 comfort indices (PMV/PPD)~\cite{iso7730}. This proxy can be replaced or enriched with feedback from wearable devices or mobile surveys~\cite{jayathissa2020humans}.

\subsection{Variable Definitions and Graph Constraints}
We denote the full variable set as $V = \{X_1, \dots, X_n\}$, partitioned into inputs, outputs, and (if present) mediators. Inputs $V_{\text{input}} = \{\text{Temperature}, \text{Humidity}, \text{AirQuality}\}$ represent controllable environmental factors, while outputs $V_{\text{output}} = \{\text{EnergyConsumption}, \\ \text{Satisfaction}\}$ capture energy and comfort outcomes. Mediators are defined as $V_{\text{mediator}} = V \setminus (V_{\text{input}} \cup V_{\text{output}})$.

To ensure physical validity, we constrain the learned DAG to be acyclic, prohibit outgoing edges from outputs, and require all edges to follow known thermodynamic and perceptual directions (e.g., environment affects comfort and energy use, not vice versa).




\subsection{Intervention Model and Constraints}
In \textsc{GRID}, interventions are modeled as controlled manipulations of input variables via building actuators (e.g., heaters, humidifiers, ventilation systems). These actuators adjust environmental conditions within safe operational bounds, enabling causal probing. For example, a temperature intervention takes the form $\text{do}(X_{\text{Temp}} = x_{\text{temp}}')$, where $x_{\text{temp}}' \in [18^\circ\text{C}, 30^\circ\text{C}]$ per ASHRAE guidelines~\cite{ashrae-energy-prediction}; similar bounds apply to humidity and air quality. Such interventions modify the joint distribution under standard do-calculus semantics, with  
\textbf{\( P(X_1, \dots, X_n \mid \text{do}(X_j = x_j')) = P(X_j = x_j') \prod_{i \ne j} P(X_i \mid \text{pa}(X_i) \setminus \{X_j\}) \)}, where ${pa}(X_i)$ refers to the parents of $(X_i)$.

\textsc{GRID} supports both fully actuated environments, where real interventions are applied, and minimally instrumented ones, where synthetic interventions are approximated using learned models. This enables broad applicability without sacrificing identifiability.

\subsection{Estimating Causal Effects}
\textsc{GRID} estimates causal effect sizes using a two-phase protocol (baseline and actuation). The effect of $X_i$ on $X_j$ is:
\[
\text{EffectSize}(X_i \rightarrow X_j) = \frac{|X_j^{\text{post}} - X_j^{\text{pre}}|}{\sigma_{\text{baseline}}(X_j)},
\]
where $\sigma_{\text{baseline}}(X_j)$ is the standard deviation of $X_j$ during baseline. Edges exceeding a threshold $\epsilon$ across trials are retained. Interventional data are prioritized over observational data due to higher causal confidence~\cite{maiti2023iccps}.

\subsection{Ground Truth Structure and Generalization}
To benchmark accuracy and robustness, we define two ground truth DAGs. The base graph encodes direct causal edges from environmental inputs to energy and satisfaction outputs:
\begin{align}
E_{\text{true}} = \{ &(\text{Temp}, \text{Energy}), (\text{Temp}, \text{Satisfaction}), \nonumber \\
&(\text{Humidity}, \text{Energy}), (\text{Humidity}, \text{Satisfaction}), \nonumber \\
&(\text{AirQuality}, \text{Energy}), (\text{AirQuality}, \text{Satisfaction}) \}
\end{align}
An extended DAG incorporates mediated dependencies such as \((\text{Temp}, \text{Humidity})\) and \((\text{Humidity}, \text{AirQuality})\), reflecting established physical couplings. While domain knowledge defines our ground truth DAGs, we acknowledge that alternate causal structures may exist and plan sensitivity analyses in future work.

\noindent
These structures support evaluation across simulated and real deployments. The bottom line is that \textsc{GRID} supports causal modeling across various instrumentation levels, building types, and occupant behaviors, enabling integration with both legacy and modern systems.
\section{\textsc{GRID} Framework Architecture} \label{arch}
Building on the formal causal discovery problem for built environments defined in the previous section, we introduce the \textsc{GRID} framework, a modular approach that integrates constraint-based algorithms, neural structural equation models, and domain-informed language models. Through an iterative process combining sensor data and controlled interventions, \textsc{GRID} constructs empirically validated DAGs representing causal relationships in building environments, enabling informed and efficient decision-making in building automation.

\begin{figure*}[!ht]
  \centering
  \includegraphics[width=\textwidth]{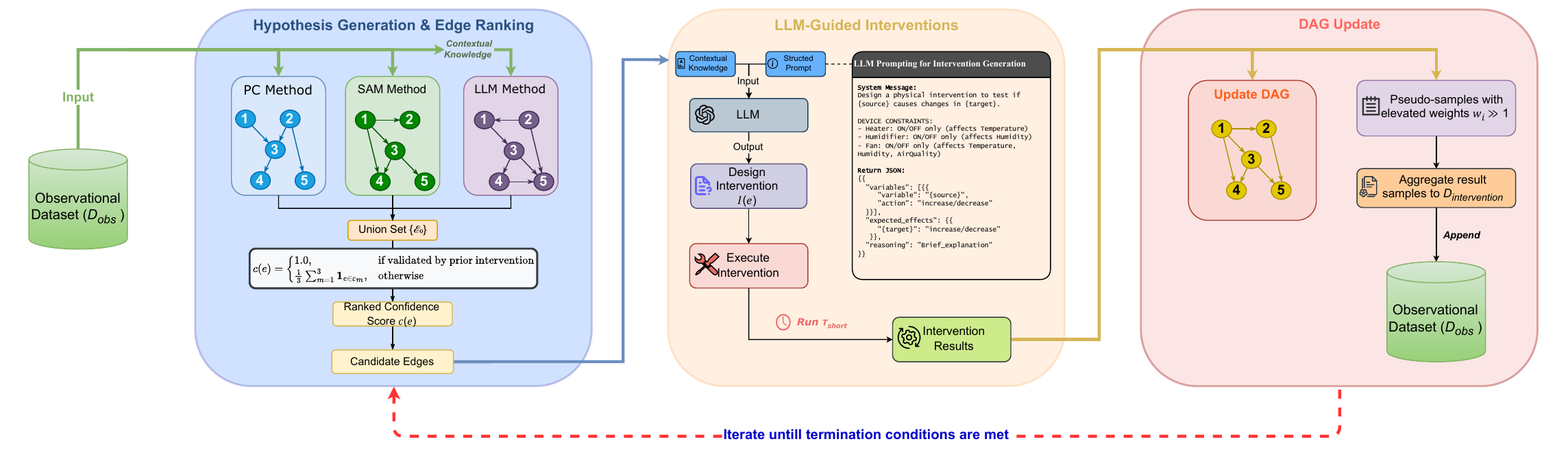}
  \caption{\textbf{System architecture of the \textsc{GRID} framework.}}
  \Description{System architecture of the \textsc{GRID} Framework.}
  \label{fig:sys_arch}
\end{figure*}


\subsection{Overview}
\textsc{GRID} follows a three-stage pipeline to generate, validate, and refine causal hypotheses in built environments. It constructs a DAG \(\mathcal{G} = (V, E)\), where \(V\) represents building variables and \(E\) candidate causal edges. The process begins by generating multiple candidate graphs \(\{\mathcal{G}_1, \mathcal{G}_2, \mathcal{G}_3\}\) using diverse methods: constraint-based PC, score-based SAM, and foundation model-guided LLM tailored to variables like temperature, humidity, and energy use. These are merged into a union graph \(\mathcal{G}_{\text{union}}\), with edges ranked by cross-method consensus. Directed edges undergo targeted interventions \(\text{do}(X_i = x_i')\) via physical actuators (e.g., HVAC), and observed responses iteratively refine the graph until convergence or stopping criteria are met. Figure~\ref{fig:sys_arch} presents the full architecture, illustrating how \textsc{GRID} combines algorithmic diversity, empirical validation, and feedback-driven refinement to support reliable causal discovery in real-world building systems.

\subsection{Multi-Method Hypothesis Generation}
Three methods generate candidate causal graphs: a constraint-based algorithm (PC), a neural network-based model (SAM), and a domain-informed large language model (LLM). Each processes observational data or domain context to capture system-level causal structure.

\paragraph{PC Algorithm: }
The Peter-Clark (PC) algorithm~\cite{spirtes2000causation} infers causal structure from observational data by testing conditional independence. It first constructs an undirected skeleton via iterative tests where \(X_i\) and \(X_j\) are conditionally independent given \(\mathbf{S}\) if \(P(X_i, X_j \mid \mathbf{S}) = P(X_i \mid \mathbf{S})P(X_j \mid \mathbf{S})\). Partial correlations are tested using the Fisher z-transform~\cite{fisher1915frequency} at significance level \(\alpha=0.05\). Edges are then oriented by standard PC rules to identify unshielded colliders and maintain acyclicity, with options for stability and collider prioritization to reduce ambiguity~\cite{spirtes2000causation}.


\paragraph{SAM Algorithm: }
The Structural Agnostic Model (SAM) infers causal structure by training a generative neural network with sparsity and acyclicity constraints~\cite{kalainathan2018structural}. It minimizes the objective \(\mathcal{L}_{\text{recon}}(\mathbf{X}, f_{\mathbf{W}}(\mathbf{X})) + \lambda_1 \|\mathbf{W}\|_1 + \lambda_2 \|\mathbf{W}\|_2^2 + \gamma \, \text{DAG}(\mathbf{W})\), where \(\mathbf{W} \in \mathbb{R}^{n \times n}\) is the weighted adjacency matrix and \(\text{DAG}(\mathbf{W})\) enforces acyclicity via a differentiable constraint. Regularization coefficients \(\lambda_1=0.1\) and \(\lambda_2=0.01\) control sparsity and overfitting, tuned for typical sensor noise. Training proceeds with learning rates \(\eta_g=0.01\) and \(\eta_d=0.005\) over 200 epochs, batch size 32, and three random restarts to improve robustness. Final edges are selected by thresholding learned weights where \(|\mathbf{W}_{ij}| > \tau\), with \(\tau\) adaptively determined from the empirical weight distribution.

\paragraph{LLM: }
\textsc{GRID} uses GPT-3.5-turbo~\cite{ye2023comprehensive} to generate causal graphs grounded in physical building principles. A fixed system message defines the model’s role as a causal discovery expert and enforces a strict JSON output format, while dynamic user messages specify variables and constraints like acyclicity and physical plausibility (see example in~\ref{llm-dag-prompt}). The prompt design here generalizes across variables in our tests, but robustness to naming conventions and unseen domains is a known limitation. The LLM infers causal hypotheses by reasoning over variable names, descriptions, and units~\cite{llmknowledgebase, kiciman2023causal}, producing a causal graph \(\mathcal{G}_{\text{LLM}}\). It is configured with temperature 1.0 and top-p 0.8 to balance creativity and domain accuracy. This graph complements other methods within \textsc{GRID}’s ensemble of hypothesis generators.

\begin{tcolorbox}[title={LLM Prompting for Base Experimental Setup}, label={llm-dag-prompt}]
\textbf{System Message:}
\begin{lstlisting}[language=Python]
You are an expert in causal discovery analyzing a smart room environment with 5 variables.
Generate a causal DAG based on physical principles and environmental systems.
Return your answer as a JSON object with this format exactly:
{"nodes": ["Temperature", "Humidity", "AirQuality", "EnergyConsumption", "OverallSatisfaction"],
 "edges": [["source", "target"], ...]}
\end{lstlisting}

\textbf{User Message:}
\begin{lstlisting}[language=Python]
Analyze this dataset with variables: Temperature, Humidity, AirQuality, EnergyConsumption, OverallSatisfaction

Rules:
1. Include directed edges based on likely causal mechanisms
2. No cycles or self-loops allowed
3. Focus on primary physical relationships
\end{lstlisting}
\end{tcolorbox}


\subsection{Edge Ranking}
After generating multiple candidate graphs, \textsc{GRID} merges the outputs into a unified representation and assigns a confidence score to each edge based on agreement across methods. Specifically, a union graph $\mathcal{G}_{\text{union}} = (V, E_{\text{union}})$ is formed by taking the union set of edges produced by the PC, SAM, and LLM generators. For each candidate edge $e_{ij}$, a confidence score $c(e_{ij})$ is computed as the proportion of methods that included the edge, i.e. edges supported by all three methods receive the highest confidence score of 1, and those proposed by only one method score 0.33. \textsc{GRID} uses these scores to prioritize intervention testing: edges with lower agreement are tested earlier to maximize information gain, while edges with full consensus are deferred unless evidence to the contrary emerges during the iterative refinement process.

\subsection{Testing via Intervention}

To validate candidate edges, \textsc{GRID} performs controlled interventions in simulation or via physical actuation, estimating the causal effect of \(X_i\) on \(X_j\) using the truncated factorization:
\[
P(X_j \mid \text{do}(X_i = x_i')) = P(X_j \mid \text{pa}(X_j) \setminus \{X_i\}, X_i = x_i'),
\]
where ${pa}(X_i)$ refers to the parents of $(X_i)$.

Interventions are generated by an LLM-guided agent that converts causal queries into device commands; for example, temperature is perturbed as \(X_{\text{Temp}} \leftarrow X_{\text{Temp}} + \delta_{\text{heat}} \cdot s_{\text{heater}}\), with similar forms for humidity and air quality, where \(s_{\text{device}} \in \{0,1\}\) is a binary control signal and \(\delta\) reflects calibrated actuation. Causal influence is measured by \(\Delta_{ij} = \mathbb{E}[X_j \mid \text{do}(X_i = x_i')] - \mathbb{E}[X_j \mid \text{do}(X_i = x_i)]\). An edge \((X_i \to X_j)\) is validated if \(|\Delta_{ij}| > \epsilon\) in at least one of \(n=3\) repeated tests, with \(\epsilon = 0.1\) as a domain-specific threshold; otherwise, it is discarded. This approach balances robustness and intervention efficiency. Our actuator model abstracts real HVAC dynamics for tractability; extending to full physical realism is left to future implementation.



\subsection{Iterative Validation Loop}
Following each intervention cycle, \textsc{GRID} updates the current graph $\mathcal{G}^{(t)}$ using the newly collected intervention data $\mathcal{D}_{\text{intervention}}^{(t)}$ and the set of validated edges $\mathcal{E}_{\text{validated}}^{(t)}$, producing an updated graph $\mathcal{G}^{(t+1)}$. Observational and interventional data are combined into a weighted dataset $\mathcal{D}_{\text{combined}} = \mathcal{D}_{\text{obs}} \cup \{(\mathbf{x}_i^{\text{int}}, w_i)\}_{i=1}^{N_{\text{int}}}$, where interventional points are assigned higher weights ($w_i = 2.0$) to reflect their stronger evidential value.

The algorithm terminates under any of the following conditions: all candidate edges have been validated ($|\mathcal{E}_{\text{validated}}| = |\mathcal{E}_{\text{union}}|$), the iteration count exceeds $T_{\max}$, or the learned graph exactly matches the ground truth ($\text{SHD}(\mathcal{G}_{\text{final}}, \mathcal{G}_{\text{true}}) = 0$). These criteria ensure that \textsc{GRID} stops either upon convergence or once sufficient causal confidence is achieved. While convergence was observed empirically, formal guarantees and iteration bounds remain open research questions.


\section{Methodology} \label{method}
To evaluate the effectiveness of the \textsc{GRID} framework, we designed a methodology such that it systematically tests and refines causal hypotheses across a range of building-relevant scenarios. This section details the intervention design, validation logic, evaluation pipeline, and benchmarking protocols.

\subsection{Edge Testing and Refinement Pipeline}
\textsc{GRID} was implemented as an iterative process that refines the learned graph $\hat{G} = (\hat{V}, \hat{E})$ over time. Each cycle includes three stages: (1) hypothesis generation and edge ranking, (2) intervention design and execution, and (3) graph updating via weighted re-learning. The evaluation procedure described here was instantiated across the six scenarios outlined in Section~\ref{exp_setup}, each reflecting a distinct challenge in building-level causal inference.

\subsubsection{Hypothesis Generation and Edge Ranking: } The first step was to generate a union set of candidate edges $\mathcal{E}_0$ by combining outputs from PC, SAM, and LLM-based methods. Each edge $e_{i \rightarrow j} \in \mathcal{E}_0$ was assigned a confidence score $c(e)$ and ranked based on cross-method agreement or prior validation:
\begin{align}
c(e) = 
\begin{cases}
1.0, & \text{if validated by prior intervention} \\
\frac{1}{3} \sum_{m=1}^{3} \mathbf{1}_{e \in \mathcal{E}_m}, & \text{otherwise}
\end{cases}
\end{align}
where $\mathcal{E}_m$ denotes the edge set produced by method $m$. Edges with $c(e) < 1.0$ were ranked in ascending order for testing.

\subsubsection{LLM-Guided Interventions: } To test low-confidence edges $e_{i \rightarrow j}$, we issued structured prompts to \texttt{gpt-3.5-turbo} to generate interventions $do(X_i = x^*)$, constrained by variable bounds and system logic. These actions, parsed as JSON $I(e)$, mimic control inputs (e.g., HVAC overrides) and were executed with retry logic and exponential backoff. Interventions ran in a short window $T_{\text{short}}$; if $|\Delta X_j| > \delta$, a longer evaluation window $T_{\text{long}}$ followed.

    


\subsubsection{Edge Validation and DAG Update: } Intervention outcomes $\{(X^{(i)}_{\text{pre}}, X^{(i)}_{\text{post}})\}_{i \in I(e)}$ were converted into pseudo-samples with elevated weights $w_i \gg 1$ and appended to the dataset. The causal discovery pipeline was then re-run over the updated dataset using weighted structure learning. Intervention samples were assigned higher weights to bias learning toward empirically supported edges.

\subsection{Evaluation Metrics} \label{eval_metrics}
Learned graphs were evaluated along two dimensions: structural fidelity to ground truth, and practical implications of false causal assumptions.

\subsubsection{Structural Accuracy: }
Structural accuracy metrics evaluated how closely the learned DAG matched the ground truth in terms of correct and incorrect edges. Given a learned DAG $\hat{G} = (\hat{V}, \hat{E})$ and ground truth $G^* = (V^*, E^*)$, true positives (TP), false positives (FP), false negatives (FN), and true negatives (TN) were computed based on edge presence. 

\subsubsection{Structural Hamming Distance (SHD): }
SHD measured the total number of edge modifications—additions, deletions, or reversals—required to convert the learned graph into the ground truth, defined as ${SHD(G^*, \hat{G}) = \sum_{i,j} |A^*_{ij} - \hat{A}_{ij}|}$, where $A^*$ and $\hat{A}$ are the adjacency matrices of $G^*$ and $\hat{G}$ respectively.

\subsubsection{Cost and Risk Metrics: }
To capture the real-world effects of model errors, we defined cost and risk metrics over false positive edges. These metrics quantify operational degradation, specifically the loss in user satisfaction and the increase in energy consumption, as measured empirically from past interventions. For a given edge $e$, the cost $\mathit{cost}(e)$ is computed as the average over all interventions $i \in I_e$ associated with that edge: $\mathit{cost}(e) = \frac{1}{|I_e|} \sum_{i \in I_e} \left( \alpha \cdot s_i + \beta \cdot \varepsilon_i \right)$, where $\alpha = \beta = 0.6$ are grid-searched weights, and $s_i$, $\varepsilon_i$ are the satisfaction loss and energy increase for intervention $i$. The satisfaction loss is $s_i = \max\left(0, \frac{S_{\text{pre},i} - S_{\text{post},i}}{S_{\text{pre},i}}\right)$ if $S_{\text{pre},i} > 0$, and $0$ otherwise. Similarly, the energy increase is $\varepsilon_i = \max\left(0, \frac{E_{\text{post},i} - E_{\text{pre},i}}{E_{\text{pre},i}}\right)$ if $E_{\text{pre},i} > 0$, and $0$ otherwise. For each method $m$, the average cost is defined over false positive edges $\hat{E}_m \setminus E^*$, where $E^*$ is the ground truth edge set and $\hat{E}_m$ the predicted edges by $m$: $\mathit{Cost}(m) = \frac{1}{|\hat{E}_m \setminus E^*|} \sum_{e \in \hat{E}_m \setminus E^*} \mathit{cost}(e)$ if $|\hat{E}_m \setminus E^*| > 0$, and $0$ otherwise.

For benchmark evaluation, cost and risk metrics were computed only for intervention-capable methods (GIES, ABCD, JCI, ICP, Causal Bandits, IID, and NOTEARS) since these require empirical intervention data. Quality filtering retained effects with $|\Delta| > 0.05$ (Cohen's medium effect threshold), and statistical confidence used t-tests with $p < 0.001 \rightarrow \text{confidence} = 0.9$.

\subsection{Benchmarking Protocol} \label{benchmark_summary}
To evaluate \textsc{GRID}, we benchmarked it against ten representative causal discovery methods spanning constraint-based, score-based, Bayesian, invariance-based, optimization-driven, and LLM-based approaches. Each was run under identical conditions with standardized datasets, iteration budgets, and evaluation metrics (Section~\ref{eval_metrics}). Implementations came from verified sources (e.g., \textit{causal-learn}, \textit{CDT}) with default or grid-tuned parameters. The suite included: \textbf{PC}~\cite{spirtes2000causation}, \textbf{SAM}~\cite{kalainathan2018structural}, \textbf{GIES}~\cite{hauser2012characterization}, \textbf{JCI}~\cite{mooij2020joint}, \textbf{ABCD}~\cite{toth2022active}, \textbf{Causal Bandits}~\cite{lattimore2016causal}, \textbf{NOTEARS-I}~\cite{zheng2020learning}, \textbf{ICP}~\cite{peters2016causal}, \textbf{IID}~\cite{zhang2023active}, and \textbf{LLM}~\cite{sunleveraging}. 

\section{Experimental Setup} \label{exp_setup}
We evaluated \textsc{GRID} against all causal discovery methods from Section~\ref{benchmark_summary} across six setups of increasing complexity, ranging from simulations to physical deployments. Each setup tested the method's robustness to observability, noise, confounding, and scale. Synthetic and dataset-driven setups (Base, Noisy, Hidden, ASHRAE) used 60 iterations; the physical setup used 15 due to hardware limits.

To compare causal complexity, we define a symbolic score \textbf{\boldmath$\mathcal{C} = n + \alpha m + \beta r + \gamma h + \delta z$}, where \(n\) is the number of observed variables, \(m\) intervenable variables, and \(r\), \(h\), \(z\) indicate presence of noise, hidden confounders, and spatial coupling. Weights \(\alpha\)-\(\delta\) capture added difficulty. This supports structured comparisons across setups (Table~\ref{tab:setup_complexity}) grounded in real-world challenges~\cite{fisher2019all,zhang2022causal,peters2017elements}.
\begin{table}[h]
\centering
\caption{\textbf{Complexity Characterization of Experimental Setups}}
\label{tab:setup_complexity}
\begin{tabular}{|l|l|l|}
\hline
\textbf{Setup} & \textbf{Complexity Expression} & \textbf{Big O Notation} \\
\hline
Base         & $n$                                 & $\mathcal{O}(n)$ \\
\hline
Noisy        & $n + \beta$                         & $\mathcal{O}(n + \beta)$ \\
\hline
Hidden       & $n + \gamma$                        & $\mathcal{O}(n + \gamma)$ \\
\hline
ASHRAE       & $n + \alpha m + \beta + \gamma$     & $\mathcal{O}(n + m)$ \\
\hline
Physical     & $n + \alpha m + \beta + \gamma + \delta$ & $\mathcal{O}(n + m + z)$ \\
\hline
Large-Scale  & $n z + \alpha m + \beta + \gamma + \delta$ & $\mathcal{O}(n z + m)$ \\
\hline
\end{tabular}
\end{table}

\vspace{7\baselineskip}
\subsection{Base Simulation (5 Variables)}
\textbf{Complexity:} $\mathcal{O}(n)$

The base setup evaluated performance under ideal conditions using a fully observable smart HVAC simulation with five variables: temperature (T), humidity (H), air quality (AQ), energy consumption (E), and occupant satisfaction (S). A ground truth DAG was manually defined based on domain knowledge. Data were generated via a custom simulator enabling batch runs and direct interventions. Energy use was estimated by nearest-neighbor interpolation over EnergyPlus data~\cite{crawley2001energyplus}, with fallbacks from ASHRAE 90.1~\cite{ashrae2019energy} and Seem’s part-load model~\cite{seem1987modeling}. Models incorporated temperature drift, ASHRAE 62.1 humidity control~\cite{ashrae2019ventilation}, and EPA-based ventilation energy~\cite{epa2009iaq}. Occupant satisfaction followed ISO 7730~\cite{iso7730}, combining PMV/PPD metrics with psychrometric inputs, and was penalized for excess energy use.

\begin{table}[H]
\centering
\small
\caption{\textbf{Base Simulation tracks temperature, humidity, and air quality (inputs) alongside energy use and occupant satisfaction (outputs); each variable has set units and ranges, and the first three causally drive the last two.}}
\begin{tabular}{lll}
\hline
\textbf{Variable Type} & \textbf{Variable} & \textbf{Range and Units} \\
\hline
\multirow{3}{*}{Input Variables} & Temperature ($T$) & 18-30°C \\
 & Humidity ($H$) & 30-70\% \\
 & Air Quality ($AQ$) & 0-500 AQI \\
\hline
\multirow{2}{*}{Output Variables} & Energy Consumption ($E$) & 0-100\% (normalized index) \\
 & Overall Satisfaction ($S$) & 0-100\% \\
\hline
\end{tabular}
\label{tab:base-variables}
\end{table}
\subsection{Noisy Simulation (5 Variables)}
\textbf{Complexity:} $\mathcal{O}(n + \beta)$

This setup extended the base by adding Gaussian noise to sensor readings to test robustness against realistic measurement uncertainty. Noise levels matched typical sensor specs: ±0.2$^\circ$C for temperature, ±2\% RH for humidity~\cite{sensirion_shtc3}, and ±15 AQI for air quality~\cite{veris_cw2}. Noise affected only observed values, keeping control variables precise to mimic real automation. The temperature range was also expanded to 18-40$^\circ$C to evaluate stability under extreme conditions.

\begin{table*}[ht]
\centering
\caption{\textbf{Observation-only performance of 11 methods across six scenarios. For each method and scenario, we report SHD (\textbf{lower is better}) and F1 score (\textbf{higher is better}).}}
\label{tab:obs_only_scenarios}
\begin{tabular}{l|cc|cc|cc|cc|cc|cc}
\toprule
\multirow{2}{*}{\textbf{Method}} & \multicolumn{2}{c|}{\textbf{Base}} & \multicolumn{2}{c|}{\textbf{Noisy}} & \multicolumn{2}{c|}{\textbf{Hidden Vars}} & \multicolumn{2}{c|}{\textbf{ASHRAE}} & \multicolumn{2}{c|}{\textbf{Physical}} & \multicolumn{2}{c}{\textbf{Large Sim}} \\
 & SHD $\downarrow$ & F1 $\uparrow$ & SHD $\downarrow$ & F1 $\uparrow$ & SHD $\downarrow$ & F1 $\uparrow$ & SHD $\downarrow$ & F1 $\uparrow$ & SHD $\downarrow$ & F1 $\uparrow$ & SHD $\downarrow$ & F1 $\uparrow$ \\
\midrule
PC & 4 & 0.60 & 4 & 0.60 & 4 & 0.60 & 4 & 0.6 & 2 & 0.88 & 49 & 0.33 \\
SAM & 8 & 0.33 & 6 & 0.25 & 8 & 0.20 & 4 & 0.33 & 7 & 0.36 & 21 & 0.16 \\
LLM & 7 & 0.36 & 7 & 0.36 & 7 & 0.36 & 4 & 0.6 & 2 & 0.86 & 17 & 0.45 \\
GIES & 4 & 0.6 & 4 & 0.60 & 4 & 0.67 & 5 & 0.29 & 5 & 0.71 & 19 & 0.34 \\
JCI & 6 & 0.40 & 6 & 0.50 & 4 & 0.67 & 2 & 0.75 & 6 & 0.50 & 48 & 0.37 \\
ABCD & 5 & 0.44 & 6 & 0.50 & 6 & 0.25 & 2 & 0.75 & 4 & 0.71 & 54 & 0.41 \\
Causal Bandits & 4 & 0.60 & 4 & 0.67 & 4 & 0.75 & 3 & 0.67 & 5 & 0.71 & 48 & 0.37 \\
ICP & 5 & 0.44 & 5 & 0.55 & 2 & 0.80 & 3 & 0.67 & 6 & 0.50 & 23 & 0.26 \\
IID & 2 & 0.8 & 6 & 0.57 & 5 & 0.62 & 2 & 0.80 & 6 & 0.50 & 28 & 0.15 \\
NOTEARS & 3 & 0.73 & 6 & 0.25 & 4 & 0.67 & 4 & 0.5 & 6 & 0.57 & 44 & 0.46 \\
\textbf{GRID-O} & 5 & 0.62 & 2 & 0.86 & 3 & 0.8 & 4 & 0.67 & 2 & 0.86 & 37 & 0.43 \\
\bottomrule
\end{tabular}
\end{table*}

\subsection{Simulation with Hidden Variables (5 Variables)}
\textbf{Complexity:} $\mathcal{O}(n + \gamma)$

This setup added latent confounders in the base setup to simulate partial observability typical in real buildings. Hidden variables such as HVAC efficiency, building envelope properties, occupancy patterns, window states, and outdoor conditions were unobserved by the algorithms but influenced observed variables and outcomes. Energy consumption and occupant satisfaction reflected time-varying effects from building physics and occupancy-driven demand, including adaptive comfort and window use.
\subsection{Real-World Dataset (ASHRAE; 5 Variables)}
\textbf{Complexity:} $\mathcal{O}(n + m)$

We used the ASHRAE Great Energy Predictor III dataset~\cite{ashrae-energy-prediction} to evaluate \textsc{GRID} and the other benchmark methods on real-world energy data from over 1,400 buildings. Six physical variables were selected: \textit{outdoor temperature, dew point, pressure, energy use, square footage,} and \textit{construction year}. Preprocessing involved daily aggregation, weather-building merging, KNN imputation, outlier removal, and robust scaling. A ground truth DAG was defined using domain expertise and physical laws. As real interventions were unavailable, we used Random Forest surrogates trained on observational data to simulate interventions and predict effects on causal children.

\subsection{Physical Deployment (5 Variables)}
\textbf{Complexity:} $\mathcal{O}(n + m + z)$

\begin{figure}[ht]
    \centering
    \includegraphics[width=\linewidth]{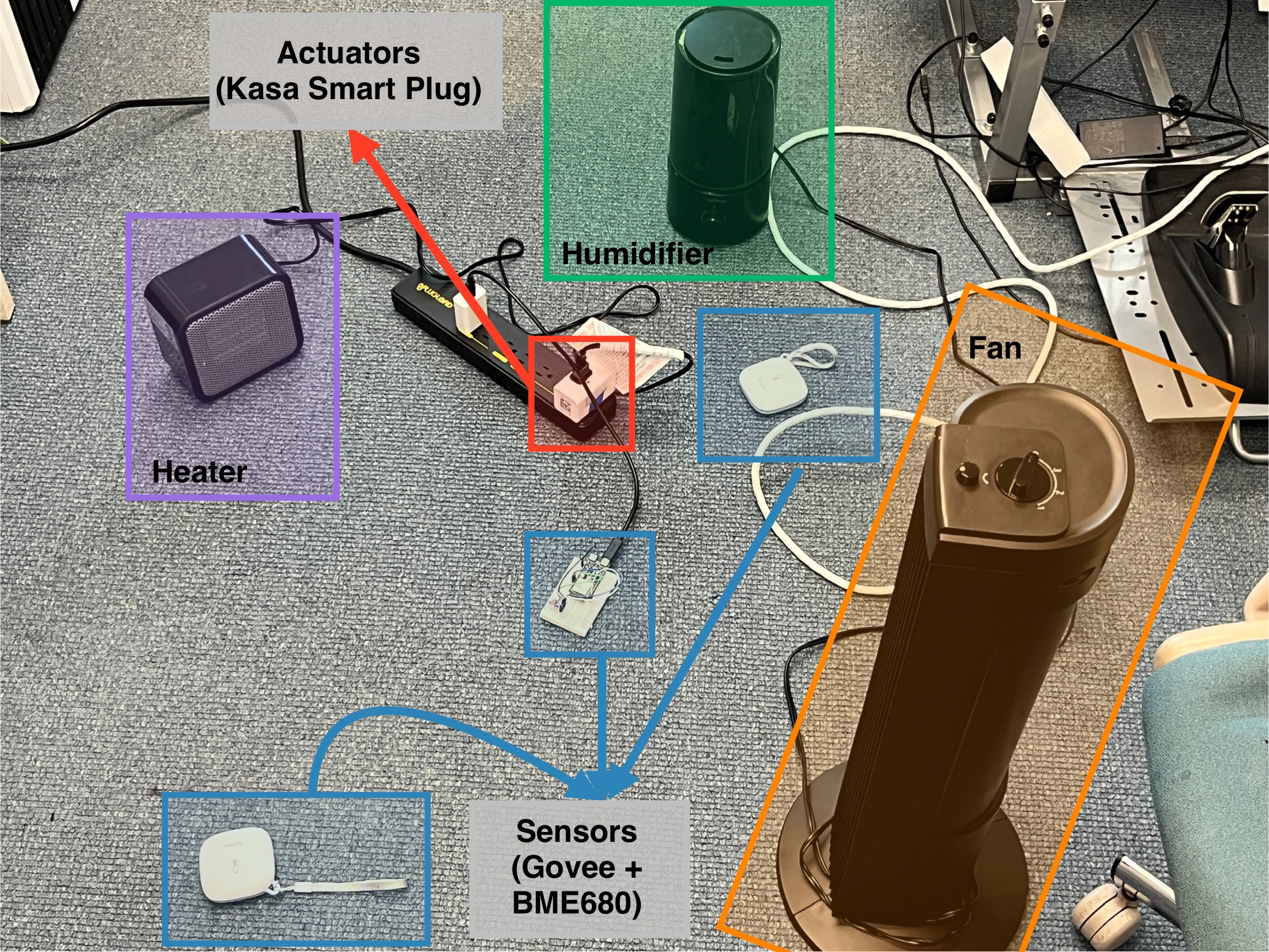}
    \caption{\textbf{\textit{Physical Deployment} setup showing sensors and actuators for environmental monitoring and control. Two Govee H5179 sensors and one BME680 track temperature, humidity, and air quality, while Kasa smart plugs control devices like a heater, fan, and humidifier. The setup supports ISO 7730 thermal comfort assessments and safe intervention timing for causal analysis.}}
    \label{fig:physical-sim}
\end{figure}

This setup validated \textsc{GRID} and the benchmark methods under real-world hardware constraints using environmental sensors, power monitors, and standardized comfort tools in a controlled space (Fig.~\ref{fig:physical-sim}).

Two Govee H5179 sensors measured temperature (±0.3°C) and humidity (±3\%), while a BME680 provided additional readings including IAQ. Three Kasa KP125M plugs monitored power with 0.1 W resolution, reporting cumulative energy usage with 1\% accuracy, and acted as actuators. PMV/PPD comfort scores followed ISO 7730~\cite{iso7730} using fixed occupant parameters. Satisfaction combined thermal and air quality metrics, following ISO 7730 standards~\cite{iso7730}. Interventions were capped at 1000/day, spaced by \(\ge 300\)s with a 600s stabilization window. Effects were quantified using Cohen’s \(d\)~\cite{cohensd}. Data were time-synced, Govee readings averaged, and energy consumption data from the Kasa plugs summed. Preprocessing used KNN imputation (\(k{=}5\)), IQR-based outlier removal, and Min-Max normalization~\cite{mining2006data}. While this setup demonstrates feasibility, scaling interventions in occupied buildings remains a practical constraint and is a direction for future deployment work.

\subsection{Large-Scale Simulation (13 Variables)}
\textbf{Complexity:} $\mathcal{O}(n z + m)$
Five inter-connected EnergyPlus zones (13 state vars each) expose control of temperature, humidity, IAQ, occupancy, HVAC set-points, and lighting. A central coordinator issues zone- and building-level interventions, captures full state, and aggregates energy, comfort, IAQ, and satisfaction metrics, reusing the single-room psychrometric and energy models while realistic schedules drive coupled dynamics for benchmarking. 

\begin{table*}[t]
    \centering
    \small
    \caption{\textbf{Precision, Recall, and F1 score for causal edge recovery across six experimental setups: \textit{Base, Noisy, Hidden, ASHRAE, Physical}, and \textit{Large-Sim}. Each method was evaluated on its ability to recover the true causal graph under varying data conditions. Bolded values indicate the highest F1 score per setup.}}
    \label{tab:core-metrics}
    \resizebox{\textwidth}{!}{%
    \begin{tabular}{l|ccc|ccc|ccc|ccc|ccc|ccc}
    \hline
    \textbf{Method} & \multicolumn{3}{c|}{\textbf{Base}} & \multicolumn{3}{c|}{\textbf{Noisy}} & \multicolumn{3}{c|}{\textbf{Hidden}} & \multicolumn{3}{c|}{\textbf{ASHRAE}} & \multicolumn{3}{c|}{\textbf{Physical}} & \multicolumn{3}{c}{\textbf{Large-Sim}} \\
     & Prec $\uparrow$ & Rec $\uparrow$ & F1 $\uparrow$ & Prec $\uparrow$ & Rec $\uparrow$ & F1 $\uparrow$ & Prec $\uparrow$ & Rec $\uparrow$ & F1 $\uparrow$ & Prec $\uparrow$ & Rec $\uparrow$ & F1 $\uparrow$ & Prec $\uparrow$ & Rec $\uparrow$ & F1 $\uparrow$ & Prec $\uparrow$ & Rec $\uparrow$ & F1 $\uparrow$ \\
    \hline
    PC   & 0.75 & 0.50 & 0.60 & 0.75 & 0.50 & 0.60 & 0.67 & 0.67 & 0.67 & 0.67 & 0.40 & 0.50 & 1.00 & 0.75 & 0.88 & 0.24 & 0.55 & 0.33 \\
    SAM  & 0.33 & 0.33 & 0.33 & 0.50 & 0.17 & 0.25 & 0.25 & 0.17 & 0.20 & 1.00 & 0.20 & 0.33 & 0.67 & 0.25 & 0.36 & 0.67 & 0.09 & 0.16 \\
    LLM  & 0.40 & 0.33 & 0.36 & 0.40 & 0.33 & 0.36 & 0.40 & 0.33 & 0.36 & 0.60 & 0.60 & 0.60 & 1.00 & 0.75 & 0.86 & 0.78 & 0.32 & 0.45 \\
    GIES & 0.50 & 0.33 & 0.40 & 0.25 & 0.33 & 0.29 & 0.38 & 0.50 & 0.43 & 0.67 & 0.40 & 0.50 &  0.80    & 1.00   & 0.89 & 0.50 & 0.45 & 0.48 \\
    JCI  & 0.40 & 0.67 & 0.50 & 0.67 & 1.00 & 0.80 & 0.38 & 0.50 & 0.43 & 0.10 & 0.20 & 0.13 & 0.75    & 0.375    & 0.50    & 0.43 & 0.86 & 0.58 \\
    ABCD & 0.67 & 0.33 & 0.44 & 0.00 & 0.00 & 0.00 & 0.50 & 0.33 & 0.40 & 0.67 & 0.40 & 0.50 &  0.50    & 0.25    & 0.34 & 0.48 & 0.73 & 0.58 \\
    Causal Bandits & 0.33 & 0.33 & 0.33 & 0.57 & 0.67 & 0.62 & 0.20 & 0.17 & 0.18 & 0.25 & 0.20 & 0.22 & 0.67    & 0.25    & 0.36    & 0.11 & 0.14 & 0.12 \\
    ICP  & 0.67 & 0.33 & 0.44 & 0.67 & 0.33 & 0.44 & 0.75 & 0.50 & 0.60 & 0.80 & 0.80 & 0.80 & 0.75    & 0.375    & 0.50    & 0.16 & 0.18 & 0.17 \\
    IID  & 0.60 & 1.00 & 0.75 & 0.60 & 1.00 & 0.75 & 0.60 & 1.00 & 0.75  & 0.27 & 0.80 & 0.40 & 0.80    & 1.00    & 0.89   & 0.28 & 1.00 & 0.44 \\
    NOTEARS & 0.25 & 0.17 & 0.20 & 0.33 & 0.50 & 0.40 & 0.71 & 0.83 & 0.77 & 0.33 & 0.20 & 0.25 & 0.83    & 0.625    & 0.714   & 0.43 & 0.59 & 0.50 \\
    \textbf{\textsc{GRID}} & 1.00 & 1.00 & \textbf{1.00} & 0.75 & 1.00 & \textbf{0.86} & 1.00 & 1.00 & \textbf{1.00} & 1.00 & 0.80 & \textbf{0.89} & 1.00 & 1.00 & \textbf{1.00} & 0.80 & 0.55 & \textbf{0.65} \\
    \hline
    \end{tabular}%
    }
\end{table*}
%

\begin{table*}[ht]
    \centering
    \caption{\textbf{Normalized intervention risk and cost for each method across six experimental setups: \textit{Base, Noisy, Hidden, ASHRAE, Physical}, and \textit{Large-Sim}. Lower values indicate fewer risks and lower costs associated with the interventions selected by each causal discovery approach. Bold values indicate the lowest risk or cost in each setup.}}
    \label{tab:risk-cost}
    \begin{tabular}{l|cc|cc|cc|cc|cc|cc}
    \hline
    \textbf{Method} 
     & \multicolumn{2}{c|}{\textbf{Base}} 
     & \multicolumn{2}{c|}{\textbf{Noisy}} 
     & \multicolumn{2}{c|}{\textbf{Hidden}} 
     & \multicolumn{2}{c|}{\textbf{ASHRAE}} 
     & \multicolumn{2}{c|}{\textbf{Physical}} 
     & \multicolumn{2}{c}{\textbf{Large-Sim}} \\
     & Risk ↓ & Cost ↓ 
     & Risk ↓ & Cost ↓ 
     & Risk ↓ & Cost ↓ 
     & Risk ↓ & Cost ↓ 
     & Risk ↓ & Cost ↓ 
     & Risk ↓ & Cost ↓ \\
    \hline
    GIES            & 0.439 & 0.472 & 0.452 & 0.486 & 0.056 & 0.06 & 0.054 & 0.067 & 0.175 & 0.218 & 0.033 & 0.046 \\
    JCI             & 0.447 & 0.480 & 0.470 & 0.505 & 0.066 & 0.071 & 0.196 & 0.245 & 0.269 & 0.336 & 0.05 & 0.1 \\
    ABCD            & 0.460 & 0.495 & 0.439 & 0.472 & 0.083 & 0.089 & 0.708 &  0.885 & 0.161 & 0.201 & 0.018 & 0.030 \\
    Causal Bandits  & 0.443 & 0.476 & 0.453 & 0.487 & 0.053 & 0.057 & 0.24 & 0.3 & 0.014 & 0.018 & 0.031 & 0.1 \\
    ICP             & 0.05 & 0.1 & 0.05 & 0.1 & 0.05 & 0.1 & 0.1 & 0.1 & 0.250 & 0.312 & 0.024 & 0.037 \\
    IID             & 0.449 & 0.483 & 0.445 & 0.479 & 0.074 & 0.080 & 0.182 & 0.227 & 0.151 & 0.188 & 0.024 & 0.037 \\
    NOTEARS         & 0.441 & 0.475 & 0.455 & 0.490 & 0.033 & 0.036 & 0.1 & 0.1 & 0.08 & 0.1 & 0.024 & 0.038 \\
    \textbf{\textsc{GRID}}            & \textbf{0.000} & \textbf{0.000} & \textbf{0.036} & \textbf{0.022} & \textbf{0.000} & \textbf{0.000} & \textbf{0.000} & \textbf{0.000} & \textbf{0.000} & \textbf{0.000} & \textbf{0.001} & \textbf{0.026} \\
    \hline
    \end{tabular}
\end{table*}

\section{Results and Discussion} \label{results}
This section presents quantitative and qualitative evaluations of \textsc{GRID} across six setups. We compare structural accuracy, intervention cost, and risk against standard baselines (Section~\ref{benchmark_summary}), a non-intervening observation-only variant, and ablated versions of our framework. Results highlight \textsc{GRID}’s effectiveness in recovering causal graphs and minimizing intervention impact across diverse conditions.

\subsection{Observation-Only Performance}
Table~\ref{tab:obs_only_scenarios} summarizes observation-only results for 11 methods across six scenarios. \textsc{GRID-O}, which applies the full \textsc{GRID} pipeline (merging PC, SAM, and LLM outputs) without intervention or iterative refinement, performs competitively across scenarios. It achieves the best or near-best SHD and F1 scores in multiple settings, including the Noisy, Hidden Variables, and Physical setups. These observation-only findings show that ensemble learning improves structure recovery from passive data but remains limited by noise, confounding, and hidden variables. Cost and risk metrics are omitted since they depend on active interventions. Subsequent sections demonstrate that full \textsc{GRID}, with targeted interventions, outperforms these observational baselines.

\subsection{Core Metrics}
Table~\ref{tab:core-metrics} reports precision, recall, and F1 scores across all six setups. \textsc{GRID} consistently achieved the highest F1 scores, including perfect recovery in \textit{Base}, \textit{Hidden}, and \textit{Physical}, and strong performance in more complex settings like \textit{Noisy} (0.86), \textit{ASHRAE} (0.89), and \textit{Large-Sim} (0.65). While GRID performs well on a 13-variable system, scaling to higher-dimensional graphs will require architectural and runtime optimization.

Performance among baselines varied. PC performed well in simpler settings (F1 = 0.88 in \textit{Physical}) but dropped under scale (\textit{Large-Sim}, 0.33). LLM peaked in \textit{Physical} (0.86) but was less consistent elsewhere. JCI showed mixed results, with moderate scores in \textit{Noisy} (0.80) and \textit{Large-Sim} (0.58), but low performance in \textit{ASHRAE} (0.13). SAM and Causal Bandits remained below 0.40 across most setups. IID and ICP had higher recall but lower precision, with F1 peaking at 0.89 (\textit{Physical}) and 0.80 (\textit{ASHRAE}), respectively. Overall, \textsc{GRID} was the most consistent and effective method across all conditions.

\subsection{Structural Accuracy}
\begin{figure*}[!ht]
\centering
\begin{minipage}{0.48\linewidth}
    \centering
    \includegraphics[width=\linewidth]{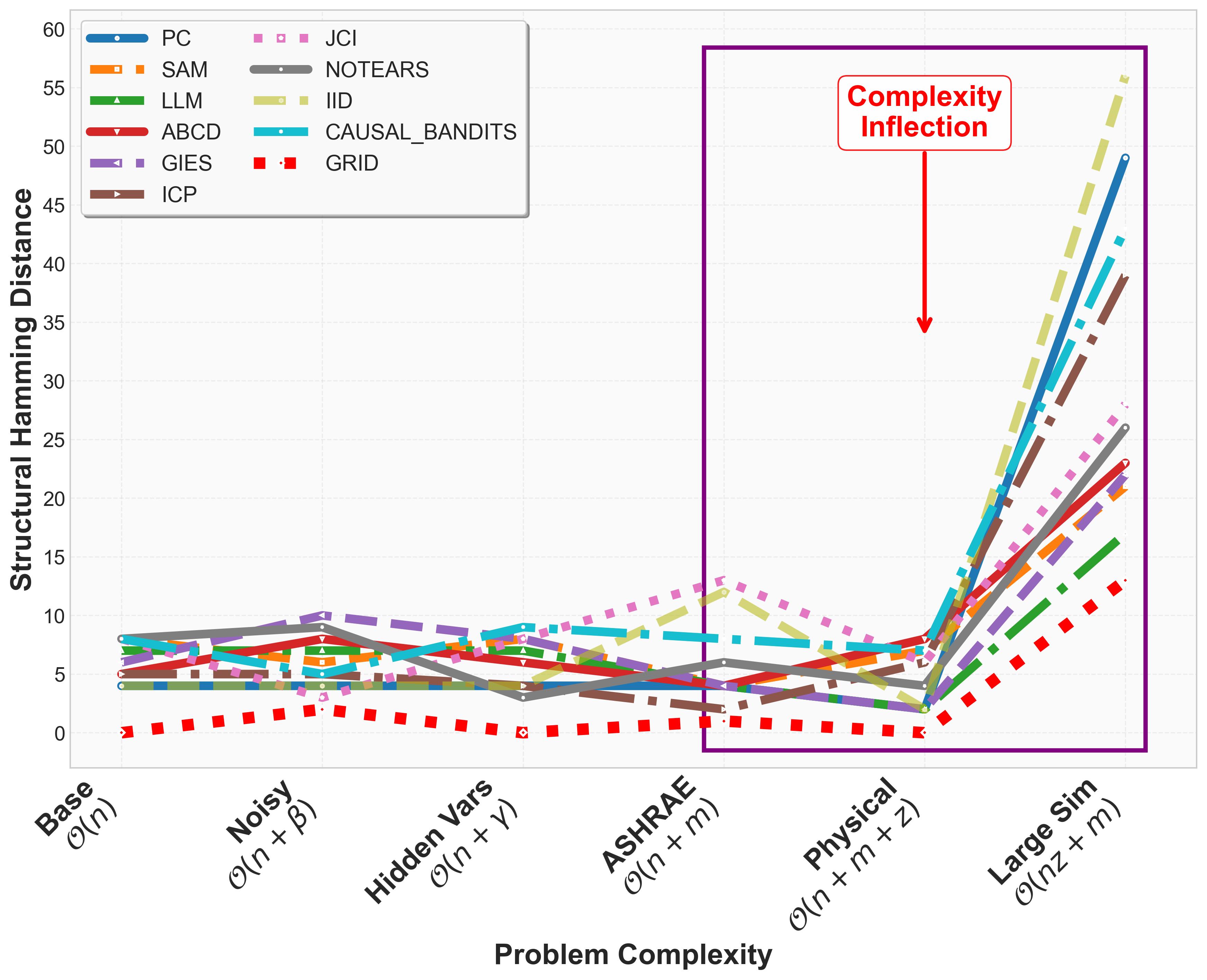}
\end{minipage}
\hspace{0.02\linewidth}
\begin{minipage}{0.38\linewidth}
\centering
    \begin{tabular}{l|c|c|c|c|c|c}
    \hline
    \textbf{Method} & \textbf{B} & \textbf{N} & \textbf{H} & \textbf{A} & \textbf{P} & \textbf{L} \\
    \hline
    PC      & 4 & 4 & 4 & 4 & 2 & 49 \\
    SAM     & 8 & 6 & 8 & 4 & 7 & 21 \\
    LLM     & 7 & 7 & 7 & 4 & 2 & 17 \\
    GIES    & 6 & 10 & 8 & 4 & 2 & 22 \\
    JCI     & 8 & 3 & 8 & 13 & 6 & 28 \\
    ABCD    & 5 & 8 & 6 & 4 & 8 & 23 \\
    Causal Bandits & 8 & 5 & 9 & 8 & 7 & 43 \\
    ICP     & 5 & 5 & 4 & 2 & 6 & 39 \\
    IID     & 4 & 4 & 4 & 12 & 2 & 56 \\
    NOTEARS & 8 & 9 & 3 & 6 & 4 & 26 \\
    \textbf{\textsc{GRID}}  & \textbf{0} & \textbf{2} & \textbf{0} & \textbf{1} & \textbf{0} & \textbf{13} \\
    \hline
    \end{tabular}

    \small
    \textbf{B}: Base, \textbf{N}: Noisy, \textbf{H}: Hidden, \\
    \textbf{A}: ASHRAE, \textbf{P}: Physical, \textbf{L}: Large-Sim
\end{minipage}
\caption{\textbf{Structural Hamming Distance (SHD) across eleven causal discovery methods and six experimental setups. Left: SHD trends over increasing problem complexity (see Section~~\ref{exp_setup}). The annotated inflection region shows a decline in structural accuracy for many methods as complexity increases. Right: SHD values for six benchmark setups. Bold values indicate lowest SHD per setup. Lower SHD means better alignment with ground truth.}}
\Description{Structural Hamming Distance (SHD) across eleven causal discovery methods and six experimental setups. Left: SHD trends over increasing problem complexity (see Section~~\ref{exp_setup}). The annotated inflection region shows a decline in structural accuracy for many methods as complexity increases. Right: SHD values for six benchmark setups. \textbf{Bold values indicate lowest SHD per setup.} Lower SHD means better alignment with ground truth.}
\label{fig:shd-summary}
\end{figure*}

Figure~\ref{fig:shd-summary} shows Structural Hamming Distance (SHD) between learned and ground truth graphs. \textsc{GRID} had the lowest SHD across all setups, with exact recovery in \textit{Base}, \textit{Hidden}, and \textit{Physical} (SHD = 0), and low error in \textit{Noisy} (2), \textit{ASHRAE} (1), and \textit{Large-Sim} (13). Baselines varied in performance. PC performed well in simpler cases (\textit{Base}: 4, \textit{Physical}: 2) but worsened under complexity (\textit{Large-Sim}: 49). IID and ABCD followed similar trends (\textit{Large-Sim}: 56 and 53). SAM, GIES, and NOTEARS showed higher SHD throughout.

SHD rose with setup complexity, especially beyond \textit{Physical}, reflecting the increasing challenge of structure recovery. As shown in Figure~\ref{fig:base-dags}, \textsc{GRID} matched the ground truth in \textit{Base}; PC was close, while IID and ABCD introduced structural errors consistent with their SHD values.


\subsection{Risk/Cost Analysis}
Table~\ref{tab:risk-cost} shows normalized intervention risk and cost across six setups. \textsc{GRID} recorded zero risk and cost in \textit{Base}, \textit{Hidden}, \textit{ASHRAE}, and \textit{Physical}, and near-zero values in \textit{Noisy} (risk = 0.036, cost = 0.022) and \textit{Large-Sim} (risk = 0.001, cost = 0.026). In comparison, baseline methods exhibited higher values across multiple setups. ABCD incurred the highest cost in \textit{ASHRAE} (0.885), and GIES showed elevated risk in \textit{Noisy} (0.452). These outcomes indicate that \textsc{GRID} consistently selected low-risk, low-cost interventions even under noise, partial observability, and deployment constraints.

\subsection{Ablation Study}
We evaluate our design choices for \textsc{GRID} using the \textit{Large-Sim} setup, which offers sufficient complexity and observability for meaningful comparison. As shown in Table~\ref{tab:ablation_components}, removing components like the edge ranker or LLM-guided interventions consistently lowers F1 and raises SHD and intervention cost. These modules serve complementary roles: ranking focuses exploration on high-confidence edges, while LLM-guided interventions reduce unnecessary effort. The \textit{minimal system} disables all intelligent components, further illustrating the performance gap and highlighting the value of each design element.
\begin{table}[H]
\centering
\footnotesize
\caption{\textbf{Ablation study on \textit{Large-Sim}. Top: hypothesis generator ablations. Middle: intervention component ablations. Bottom: minimal system vs. full \textsc{GRID}.}}
\label{tab:ablation_components}
\begin{tabular}{@{}l|ccc|c|cc@{}}
\toprule
Configuration & Prec. $\uparrow$ & Rec. $\uparrow$ & F1 $\uparrow$ & SHD $\downarrow$ & Cost $\downarrow$ & Risk $\downarrow$ \\
\midrule
PC + LLM       & 0.778 & 0.318 & 0.452 & 17 & 0.222 & 0.049 \\
PC + SAM       & 0.500 & 0.318 & 0.389 & 22 & 0.500 & 0.250 \\
PC only        & 0.667 & 0.273 & 0.387 & 19 & 0.333 & 0.111 \\
LLM only       & 0.800 & 0.182 & 0.296 & 19 & 0.200 & 0.040 \\
SAM + LLM      & 0.500 & 0.136 & 0.214 & 22 & 0.500 & 0.250 \\
SAM only       & 0.000 & 0.000 & 0.000 & 24 & 1.000 & 1.000 \\
\midrule
No edge ranking      & 0.609 & 0.636 & 0.622 & 17 & 0.391 & 0.153 \\
No LLM interventions & 0.522 & 0.545 & 0.533 & 21 & 0.478 & 0.229 \\
No edge validation    & 0.579 & 0.500 & 0.537 & 19 & 0.421 & 0.177 \\
No dataset update       & 0.636 & 0.318 & 0.424 & 19 & 0.364 & 0.132 \\
No ranking + intervention     & 0.571 & 0.545 & 0.558 & 19 & 0.429 & 0.184 \\
No intervention + update   & 0.684 & 0.591 & 0.634 & 15 & 0.316 & 0.100 \\
\midrule
Minimal system & 0.583 & 0.318 & 0.412 & 20 & 0.417 & 0.174\\
\textbf{GRID (full)} & \textbf{0.800} & \textbf{0.550} & \textbf{0.650} & \textbf{13} & \textbf{0.026} & \textbf{0.001} \\
\bottomrule
\end{tabular}
\end{table}
\noindent
The results show that while \textsc{GRID-O} offers reliable passive estimates, active interventions are crucial under noise and confounding. \textsc{GRID} achieves lower structural error and intervention cost than baselines as system complexity grows, demonstrating the benefit of iterative, intervention-guided refinement. In more instrumented environments~\cite{alsafery2023sensing, heidary2023smart}, as well as legacy setups with limited data~\cite{buildings13123088, ntafalias2024smart}, recovered graphs capture interpretable dependencies that support efficient interventions and align with evolving building standards~\cite{DOE2021}, without system-specific tuning. A detailed taxonomy of \textsc{GRID}'s failure cases is beyond this work's scope but is part of ongoing evaluation efforts.
\section{Conclusion} \label{fin}
\textbf{GRID} synergizes programmatic interventions, edge-ranking, and scalable graph search to expose causal structure in noisy, partially observed buildings, from sensor-sparse legacy systems to modern sensor-rich deployments. Across six setups—simulated, testbed, and production datasets—it outperforms classical, interventional, and neural baselines in SHD and F1 score, delivering interpretable graphs that enable precise HVAC control, efficiency gains, and fault diagnostics.
% ...

\bibliographystyle{ACM-Reference-Format}
\bibliography{refs}


\begin{thebibliography}{63}


\ifx \showCODEN    \undefined \def \showCODEN     #1{\unskip}     \fi
\ifx \showISBNx    \undefined \def \showISBNx     #1{\unskip}     \fi
\ifx \showISBNxiii \undefined \def \showISBNxiii  #1{\unskip}     \fi
\ifx \showISSN     \undefined \def \showISSN      #1{\unskip}     \fi
\ifx \showLCCN     \undefined \def \showLCCN      #1{\unskip}     \fi
\ifx \shownote     \undefined \def \shownote      #1{#1}          \fi
\ifx \showarticletitle \undefined \def \showarticletitle #1{#1}   \fi
\ifx \showURL      \undefined \def \showURL       {\relax}        \fi
\providecommand\bibfield[2]{#2}
\providecommand\bibinfo[2]{#2}
\providecommand\natexlab[1]{#1}
\providecommand\showeprint[2][]{arXiv:#2}

\bibitem[iea(2019)]%
        {iea2019global}
 \bibinfo{year}{2019}\natexlab{}.
\newblock \bibinfo{booktitle}{\emph{Global Status Report for Buildings and Construction 2019: Towards a Zero‑Emission, Efficient and Resilient Buildings and Construction Sector}}.
\newblock \bibinfo{type}{{T}echnical {R}eport}. \bibinfo{institution}{International Energy Agency and UN Environment Programme}, \bibinfo{address}{Paris}.
\newblock
\urldef\tempurl%
\url{https://www.iea.org/reports/global-status-report-for-buildings-and-construction-2019}
\showURL{%
\tempurl}


\bibitem[AG(2022)]%
        {sensirion_shtc3}
\bibfield{author}{\bibinfo{person}{Sensirion AG}.} \bibinfo{year}{2022}\natexlab{}.
\newblock \bibinfo{title}{SHTC3 – Digital Temperature and Humidity Sensor}.
\newblock \bibinfo{howpublished}{\url{https://sensirion.com/products/catalog/SHTC3}}.
\newblock
\newblock
\shownote{Typical accuracy: plus/minus 0.2 degrees C temperature; plus/minus 2 percent RH humidity}.


\bibitem[Alsafery et~al\mbox{.}(2023)]%
        {alsafery2023sensing}
\bibfield{author}{\bibinfo{person}{Wael Alsafery}, \bibinfo{person}{Omer Rana}, {and} \bibinfo{person}{Charith Perera}.} \bibinfo{year}{2023}\natexlab{}.
\newblock \showarticletitle{Sensing within smart buildings: A survey}.
\newblock \bibinfo{journal}{\emph{Comput. Surveys}} \bibinfo{volume}{55}, \bibinfo{number}{13s} (\bibinfo{year}{2023}), \bibinfo{pages}{1--35}.
\newblock


\bibitem[Ashrae(1989)]%
        {ashrae2019ventilation}
\bibfield{author}{\bibinfo{person}{Ashrae}.} \bibinfo{year}{1989}\natexlab{}.
\newblock \bibinfo{booktitle}{\emph{ASHRAE Standard 62-1989: Ventilation for Acceptable Indoor Air Quality}}.
\newblock Atlanta, GA.
\newblock


\bibitem[Ashrae(2025)]%
        {ashrae40}
\bibfield{author}{\bibinfo{person}{Ashrae}.} \bibinfo{year}{2025}\natexlab{}.
\newblock \bibinfo{title}{ANSI/ASHRAE Standard 40-2025: Method of Testing for Rating Heat-operated Unitary Air-conditioning and Heat Pump Equipment}.
\newblock


\bibitem[Burman et~al\mbox{.}(2014)]%
        {burman2014review}
\bibfield{author}{\bibinfo{person}{Emilie Burman}, \bibinfo{person}{Dejan Mumovic}, {and} \bibinfo{person}{John Kimpian}.} \bibinfo{year}{2014}\natexlab{}.
\newblock \showarticletitle{A comparative study of benchmarking approaches for non-domestic buildings: Part 2 – Bottom-up approach}.
\newblock \bibinfo{journal}{\emph{Building Services Engineering Research and Technology}} \bibinfo{volume}{35}, \bibinfo{number}{4} (\bibinfo{year}{2014}), \bibinfo{pages}{407--428}.
\newblock


\bibitem[Chen et~al\mbox{.}(2022a)]%
        {chen2022design}
\bibfield{author}{\bibinfo{person}{Xia Chen}, \bibinfo{person}{Jimmy Abualdenien}, \bibinfo{person}{Manav~Mahan Singh}, \bibinfo{person}{André Borrmann}, {and} \bibinfo{person}{Philipp Geyer}.} \bibinfo{year}{2022}\natexlab{a}.
\newblock \showarticletitle{Introducing causal inference in the energy-efficient building design process}.
\newblock \bibinfo{journal}{\emph{Energy and Buildings}}  \bibinfo{volume}{277} (\bibinfo{year}{2022}), \bibinfo{pages}{112583}.
\newblock
\href{https://doi.org/10.1016/j.enbuild.2022.112583}{doi:\nolinkurl{10.1016/j.enbuild.2022.112583}}


\bibitem[Chen et~al\mbox{.}(2022b)]%
        {chen2022introducing}
\bibfield{author}{\bibinfo{person}{Xia Chen}, \bibinfo{person}{Jimmy Abualdenien}, \bibinfo{person}{Manav~Mahan Singh}, \bibinfo{person}{Andr{\'e} Borrmann}, {and} \bibinfo{person}{Philipp Geyer}.} \bibinfo{year}{2022}\natexlab{b}.
\newblock \showarticletitle{Introducing causal inference in the energy-efficient building design process}.
\newblock \bibinfo{journal}{\emph{Energy and Buildings}}  \bibinfo{volume}{277} (\bibinfo{year}{2022}), \bibinfo{pages}{112583}.
\newblock


\bibitem[Colombo et~al\mbox{.}(2012)]%
        {colombo2012learning}
\bibfield{author}{\bibinfo{person}{Diego Colombo}, \bibinfo{person}{Marloes~H Maathuis}, \bibinfo{person}{Markus Kalisch}, {and} \bibinfo{person}{Thomas~S Richardson}.} \bibinfo{year}{2012}\natexlab{}.
\newblock \showarticletitle{Learning high-dimensional directed acyclic graphs with latent and selection variables}.
\newblock \bibinfo{journal}{\emph{The Annals of Statistics}} (\bibinfo{year}{2012}), \bibinfo{pages}{294--321}.
\newblock


\bibitem[Crawley et~al\mbox{.}(2001)]%
        {crawley2001energyplus}
\bibfield{author}{\bibinfo{person}{Drury~B Crawley}, \bibinfo{person}{Linda~K Lawrie}, \bibinfo{person}{Frederick~C Winkelmann}, \bibinfo{person}{Walter~F Buhl}, \bibinfo{person}{Y~Joe Huang}, \bibinfo{person}{Curtis~O Pedersen}, \bibinfo{person}{Richard~K Strand}, \bibinfo{person}{Richard~J Liesen}, \bibinfo{person}{Daniel~E Fisher}, \bibinfo{person}{Michael~J Witte}, {et~al\mbox{.}}} \bibinfo{year}{2001}\natexlab{}.
\newblock \showarticletitle{EnergyPlus: creating a new-generation building energy simulation program}.
\newblock \bibinfo{journal}{\emph{Energy and buildings}} \bibinfo{volume}{33}, \bibinfo{number}{4} (\bibinfo{year}{2001}), \bibinfo{pages}{319--331}.
\newblock


\bibitem[Czekster et~al\mbox{.}(2022)]%
        {czekster2022incorporating}
\bibfield{author}{\bibinfo{person}{Ricardo~M Czekster}, \bibinfo{person}{Roberto Metere}, {and} \bibinfo{person}{Charles Morisset}.} \bibinfo{year}{2022}\natexlab{}.
\newblock \showarticletitle{Incorporating cyber threat intelligence into complex cyber-physical systems: A STIX model for active buildings}.
\newblock \bibinfo{journal}{\emph{Applied Sciences}} \bibinfo{volume}{12}, \bibinfo{number}{10} (\bibinfo{year}{2022}), \bibinfo{pages}{5005}.
\newblock


\bibitem[de~Wilde(2011)]%
        {de2011closing}
\bibfield{author}{\bibinfo{person}{Pieter de Wilde}.} \bibinfo{year}{2011}\natexlab{}.
\newblock \showarticletitle{Closing the loop–Benchmarking energy performance of buildings}.
\newblock \bibinfo{journal}{\emph{Building Research \& Information}} \bibinfo{volume}{39}, \bibinfo{number}{4} (\bibinfo{year}{2011}), \bibinfo{pages}{388--400}.
\newblock


\bibitem[Diener(2010)]%
        {cohensd}
\bibfield{author}{\bibinfo{person}{Marc~J Diener}.} \bibinfo{year}{2010}\natexlab{}.
\newblock \showarticletitle{Cohen's d}.
\newblock \bibinfo{journal}{\emph{The Corsini encyclopedia of psychology}} (\bibinfo{year}{2010}), \bibinfo{pages}{1--1}.
\newblock


\bibitem[Fisher et~al\mbox{.}(2019)]%
        {fisher2019all}
\bibfield{author}{\bibinfo{person}{Aaron Fisher}, \bibinfo{person}{Cynthia Rudin}, {and} \bibinfo{person}{Francesca Dominici}.} \bibinfo{year}{2019}\natexlab{}.
\newblock \showarticletitle{All models are wrong, but many are useful: Learning a variable's importance by studying an entire class of prediction models simultaneously}.
\newblock \bibinfo{journal}{\emph{Journal of Machine Learning Research}} \bibinfo{volume}{20}, \bibinfo{number}{177} (\bibinfo{year}{2019}), \bibinfo{pages}{1--81}.
\newblock


\bibitem[Fisher(1915)]%
        {fisher1915frequency}
\bibfield{author}{\bibinfo{person}{Ronald~A Fisher}.} \bibinfo{year}{1915}\natexlab{}.
\newblock \showarticletitle{Frequency distribution of the values of the correlation coefficient in samples from an indefinitely large population}.
\newblock \bibinfo{journal}{\emph{Biometrika}} \bibinfo{volume}{10}, \bibinfo{number}{4} (\bibinfo{year}{1915}), \bibinfo{pages}{507--521}.
\newblock


\bibitem[Glymour et~al\mbox{.}(2019)]%
        {glymour2019review}
\bibfield{author}{\bibinfo{person}{Clark Glymour}, \bibinfo{person}{Kun Zhang}, {and} \bibinfo{person}{Peter Spirtes}.} \bibinfo{year}{2019}\natexlab{}.
\newblock \showarticletitle{Review of causal discovery methods based on graphical models}.
\newblock \bibinfo{journal}{\emph{Frontiers in genetics}}  \bibinfo{volume}{10} (\bibinfo{year}{2019}), \bibinfo{pages}{524}.
\newblock


\bibitem[Gümürçinler and Akboğa-Kale(2023)]%
        {buildings13123088}
\bibfield{author}{\bibinfo{person}{Tuğçe Gümürçinler} {and} \bibinfo{person}{Özge Akboğa-Kale}.} \bibinfo{year}{2023}\natexlab{}.
\newblock \showarticletitle{Evaluation of Occupational Safety in Restoration Projects of Historic Buildings: Risk Analysis with Selected Projects}.
\newblock \bibinfo{journal}{\emph{Buildings}} \bibinfo{volume}{13}, \bibinfo{number}{12} (\bibinfo{year}{2023}).
\newblock
\showISSN{2075-5309}
\href{https://doi.org/10.3390/buildings13123088}{doi:\nolinkurl{10.3390/buildings13123088}}


\bibitem[Hauser and B{\"u}hlmann(2012)]%
        {hauser2012characterization}
\bibfield{author}{\bibinfo{person}{Alain Hauser} {and} \bibinfo{person}{Peter B{\"u}hlmann}.} \bibinfo{year}{2012}\natexlab{}.
\newblock \showarticletitle{Characterization and greedy learning of interventional Markov equivalence classes of directed acyclic graphs}.
\newblock \bibinfo{journal}{\emph{The Journal of Machine Learning Research}} \bibinfo{volume}{13}, \bibinfo{number}{1} (\bibinfo{year}{2012}), \bibinfo{pages}{2409--2464}.
\newblock


\bibitem[Heidary et~al\mbox{.}(2023)]%
        {heidary2023smart}
\bibfield{author}{\bibinfo{person}{Rooh Heidary}, \bibinfo{person}{Jubilee Rao}, {and} \bibinfo{person}{Olivia Pinon~Fischer}.} \bibinfo{year}{2023}\natexlab{}.
\newblock \bibinfo{booktitle}{\emph{Smart Buildings in the IoT Era – Necessity, Challenges, and Opportunities}}.
\newblock \bibinfo{pages}{1--21}.
\newblock
\showISBNx{978-3-030-72322-4}
\href{https://doi.org/10.1007/978-3-030-72322-4_115-1}{doi:\nolinkurl{10.1007/978-3-030-72322-4_115-1}}


\bibitem[Howard et~al\mbox{.}(2019)]%
        {ashrae-energy-prediction}
\bibfield{author}{\bibinfo{person}{Addison Howard}, \bibinfo{person}{Chris Balbach}, \bibinfo{person}{Clayton Miller}, \bibinfo{person}{Jeff Haberl}, \bibinfo{person}{Krishnan Gowri}, {and} \bibinfo{person}{Sohier Dane}.} \bibinfo{year}{2019}\natexlab{}.
\newblock \bibinfo{title}{ASHRAE - Great Energy Predictor III}.
\newblock
\urldef\tempurl%
\url{https://kaggle.com/competitions/ashrae-energy-prediction}
\showURL{%
\tempurl}
\newblock
\shownote{Kaggle}.


\bibitem[Industries(2020)]%
        {veris_cw2}
\bibfield{author}{\bibinfo{person}{Veris Industries}.} \bibinfo{year}{2020}\natexlab{}.
\newblock \bibinfo{title}{Veris CW2XP2AV Air Quality Sensor Datasheet}.
\newblock \bibinfo{howpublished}{\url{https://www.veris.com/}}.
\newblock
\newblock
\shownote{VOC accuracy: plus/minus 15 percent (analogous to plus/minus 15 AQI noise level)}.


\bibitem[{International Organization for Standardization}(2005)]%
        {iso7730}
\bibfield{author}{\bibinfo{person}{{International Organization for Standardization}}.} \bibinfo{year}{2005}\natexlab{}.
\newblock \showarticletitle{ISO 7730: Ergonomics of the thermal environment}.
\newblock \bibinfo{journal}{\emph{International Standard}}  \bibinfo{volume}{7730} (\bibinfo{year}{2005}), \bibinfo{pages}{1--52}.
\newblock


\bibitem[Jayathissa et~al\mbox{.}(2020)]%
        {jayathissa2020humans}
\bibfield{author}{\bibinfo{person}{Prageeth Jayathissa}, \bibinfo{person}{Matias Quintana}, \bibinfo{person}{Mahmoud Abdelrahman}, {and} \bibinfo{person}{Clayton Miller}.} \bibinfo{year}{2020}\natexlab{}.
\newblock \showarticletitle{Humans-as-a-sensor for buildings—intensive longitudinal indoor comfort models}.
\newblock \bibinfo{journal}{\emph{Buildings}} \bibinfo{volume}{10}, \bibinfo{number}{10} (\bibinfo{year}{2020}), \bibinfo{pages}{174}.
\newblock


\bibitem[Kalainathan et~al\mbox{.}(2018)]%
        {kalainathan2018structural}
\bibfield{author}{\bibinfo{person}{Diviyan Kalainathan}, \bibinfo{person}{Olivier Goudet}, \bibinfo{person}{Isabelle Guyon}, \bibinfo{person}{David Lopez-Paz}, {and} \bibinfo{person}{Mich{\`e}le Sebag}.} \bibinfo{year}{2018}\natexlab{}.
\newblock \showarticletitle{Structural Agnostic Modeling: Adversarial Learning of Causal Graphs}.
\newblock \bibinfo{journal}{\emph{arXiv preprint arXiv:1803.04929}} (\bibinfo{year}{2018}).
\newblock


\bibitem[Kathirgamanathan et~al\mbox{.}(2021)]%
        {kathirgamanathan2021data}
\bibfield{author}{\bibinfo{person}{Anjukan Kathirgamanathan}, \bibinfo{person}{Mattia De~Rosa}, \bibinfo{person}{Eleni Mangina}, {and} \bibinfo{person}{Donal~P Finn}.} \bibinfo{year}{2021}\natexlab{}.
\newblock \showarticletitle{Data-driven predictive control for unlocking building energy flexibility: A review}.
\newblock \bibinfo{journal}{\emph{Renewable and Sustainable Energy Reviews}}  \bibinfo{volume}{135} (\bibinfo{year}{2021}), \bibinfo{pages}{110120}.
\newblock


\bibitem[Khatibi et~al\mbox{.}(2024)]%
        {khatibi2024alcm}
\bibfield{author}{\bibinfo{person}{Elahe Khatibi}, \bibinfo{person}{Mahyar Abbasian}, \bibinfo{person}{Zhongqi Yang}, \bibinfo{person}{Iman Azimi}, {and} \bibinfo{person}{Amir~M.\ Rahmani}.} \bibinfo{year}{2024}\natexlab{}.
\newblock \showarticletitle{ALCM: Autonomous LLM-Augmented Causal Discovery Framework}.
\newblock \bibinfo{journal}{\emph{arXiv preprint arXiv:2405.01744}} (\bibinfo{year}{2024}).
\newblock


\bibitem[K{\i}c{\i}man et~al\mbox{.}(2023)]%
        {kiciman2023causal}
\bibfield{author}{\bibinfo{person}{Emre K{\i}c{\i}man}, \bibinfo{person}{Robert Ness}, \bibinfo{person}{Amit Sharma}, {and} \bibinfo{person}{Chenhao Tan}.} \bibinfo{year}{2023}\natexlab{}.
\newblock \showarticletitle{Causal reasoning and large language models: Opening a new frontier for causality. arXiv}.
\newblock \bibinfo{journal}{\emph{arXiv preprint arXiv:2305.00050}} (\bibinfo{year}{2023}).
\newblock


\bibitem[Klein(2000)]%
        {klein2000trnsys}
\bibfield{author}{\bibinfo{person}{Sanford~A Klein}.} \bibinfo{year}{2000}\natexlab{}.
\newblock \bibinfo{title}{TRNSYS 16: A transient system simulation program}.
\newblock
\newblock
\shownote{Solar Energy Laboratory, University of Wisconsin-Madison}.


\bibitem[Kleissl and Agarwal(2010)]%
        {5523244}
\bibfield{author}{\bibinfo{person}{Jan Kleissl} {and} \bibinfo{person}{Yuvraj Agarwal}.} \bibinfo{year}{2010}\natexlab{}.
\newblock \showarticletitle{Cyber-physical energy systems: Focus on smart buildings}. In \bibinfo{booktitle}{\emph{Design Automation Conference}}. \bibinfo{pages}{749--754}.
\newblock
\href{https://doi.org/10.1145/1837274.1837464}{doi:\nolinkurl{10.1145/1837274.1837464}}


\bibitem[Lattimore et~al\mbox{.}(2016)]%
        {lattimore2016causal}
\bibfield{author}{\bibinfo{person}{Finnian Lattimore}, \bibinfo{person}{Tor Lattimore}, {and} \bibinfo{person}{Mark~D Reid}.} \bibinfo{year}{2016}\natexlab{}.
\newblock \showarticletitle{Causal bandits: Learning good interventions via causal inference}.
\newblock \bibinfo{journal}{\emph{Advances in neural information processing systems}}  \bibinfo{volume}{29} (\bibinfo{year}{2016}).
\newblock


\bibitem[Long et~al\mbox{.}(2023)]%
        {long2023causal}
\bibfield{author}{\bibinfo{person}{Stephanie Long}, \bibinfo{person}{Alexandre Pich{\'e}}, \bibinfo{person}{Valentina Zantedeschi}, \bibinfo{person}{Tibor Schuster}, {and} \bibinfo{person}{Alexandre Drouin}.} \bibinfo{year}{2023}\natexlab{}.
\newblock \showarticletitle{Causal discovery with language models as imperfect experts}.
\newblock \bibinfo{journal}{\emph{arXiv preprint arXiv:2307.02390}} (\bibinfo{year}{2023}).
\newblock


\bibitem[Maiti et~al\mbox{.}(2023)]%
        {maiti2023iccps}
\bibfield{author}{\bibinfo{person}{Rajib~Ranjan Maiti}, \bibinfo{person}{Sridhar Adepu}, {and} \bibinfo{person}{Emil Lupu}.} \bibinfo{year}{2023}\natexlab{}.
\newblock \showarticletitle{ICCPS: Impact discovery using causal inference for cyber attacks in CPSs}.
\newblock \bibinfo{journal}{\emph{arXiv preprint arXiv:2307.14161}} (\bibinfo{year}{2023}).
\newblock


\bibitem[Meinshausen et~al\mbox{.}(2016)]%
        {meinshausen2016methods}
\bibfield{author}{\bibinfo{person}{Nicolai Meinshausen}, \bibinfo{person}{Alain Hauser}, \bibinfo{person}{Joris~M Mooij}, \bibinfo{person}{Jonas Peters}, \bibinfo{person}{Philip Versteeg}, {and} \bibinfo{person}{Peter B{\"u}hlmann}.} \bibinfo{year}{2016}\natexlab{}.
\newblock \showarticletitle{Methods for causal inference from gene perturbation experiments and validation}.
\newblock \bibinfo{journal}{\emph{Proceedings of the National Academy of Sciences}} \bibinfo{volume}{113}, \bibinfo{number}{27} (\bibinfo{year}{2016}), \bibinfo{pages}{7361--7368}.
\newblock


\bibitem[Mining(2006)]%
        {mining2006data}
\bibfield{author}{\bibinfo{person}{What Is~Data Mining}.} \bibinfo{year}{2006}\natexlab{}.
\newblock \showarticletitle{Data mining: Concepts and techniques}.
\newblock \bibinfo{journal}{\emph{Morgan Kaufinann}} \bibinfo{volume}{10}, \bibinfo{number}{559-569} (\bibinfo{year}{2006}), \bibinfo{pages}{4}.
\newblock


\bibitem[Monti et~al\mbox{.}(2020)]%
        {monti2020causal}
\bibfield{author}{\bibinfo{person}{Ricardo~Pio Monti}, \bibinfo{person}{Kun Zhang}, {and} \bibinfo{person}{Aapo Hyv{\"a}rinen}.} \bibinfo{year}{2020}\natexlab{}.
\newblock \showarticletitle{Causal discovery with general non-linear relationships using non-linear ICA}. In \bibinfo{booktitle}{\emph{Uncertainty in artificial intelligence}}. PMLR, \bibinfo{pages}{186--195}.
\newblock


\bibitem[Mooij et~al\mbox{.}(2020)]%
        {mooij2020joint}
\bibfield{author}{\bibinfo{person}{Joris~M Mooij}, \bibinfo{person}{Sara Magliacane}, {and} \bibinfo{person}{Tom Claassen}.} \bibinfo{year}{2020}\natexlab{}.
\newblock \showarticletitle{Joint causal inference from multiple contexts}.
\newblock \bibinfo{journal}{\emph{Journal of machine learning research}} \bibinfo{volume}{21}, \bibinfo{number}{99} (\bibinfo{year}{2020}), \bibinfo{pages}{1--108}.
\newblock


\bibitem[Nambiar et~al\mbox{.}(2023)]%
        {ashrae2019energy}
\bibfield{author}{\bibinfo{person}{Chitra Nambiar}, \bibinfo{person}{Michael Rosenberg}, {and} \bibinfo{person}{Samuel Rosenberg}.} \bibinfo{year}{2023}\natexlab{}.
\newblock \showarticletitle{End Use Analysis Of ANSI/ASHRAE/IES Standard 90.1-2019}.
\newblock \bibinfo{journal}{\emph{ASHRAE Journal}} \bibinfo{volume}{65}, \bibinfo{number}{4} (\bibinfo{year}{2023}), \bibinfo{pages}{34--42}.
\newblock


\bibitem[Neogi et~al\mbox{.}(2025)]%
        {neogi2025insightbuild}
\bibfield{author}{\bibinfo{person}{Pinaki Prasad~Guha Neogi}, \bibinfo{person}{Ahmad Mohammadshirazi}, {and} \bibinfo{person}{Rajiv Ramnath}.} \bibinfo{year}{2025}\natexlab{}.
\newblock \showarticletitle{InsightBuild: LLM-Powered Causal Reasoning in Smart Building Systems}. In \bibinfo{booktitle}{\emph{ICML 2025 CO-BUILD Workshop on Computational Optimization of Buildings}}. \bibinfo{publisher}{PMLR}, \bibinfo{address}{Vienna, Austria}.
\newblock


\bibitem[Ntafalias et~al\mbox{.}(2024)]%
        {ntafalias2024smart}
\bibfield{author}{\bibinfo{person}{Aristotelis Ntafalias}, \bibinfo{person}{Panagiotis Papadopoulos}, \bibinfo{person}{Alfonso~P Ramallo-Gonz{\'a}lez}, \bibinfo{person}{Antonio~F Skarmeta-G{\'o}mez}, \bibinfo{person}{Juan S{\'a}nchez-Valverde}, \bibinfo{person}{Maria~C Vlachou}, \bibinfo{person}{Rafael Mar{\'\i}n-P{\'e}rez}, \bibinfo{person}{Alfredo Quesada-S{\'a}nchez}, \bibinfo{person}{Fergal Purcell}, {and} \bibinfo{person}{Stephen Wright}.} \bibinfo{year}{2024}\natexlab{}.
\newblock \showarticletitle{Smart buildings with legacy equipment: A case study on energy savings and cost reduction through an IoT platform in Ireland and Greece}.
\newblock \bibinfo{journal}{\emph{Results In Engineering}}  \bibinfo{volume}{22} (\bibinfo{year}{2024}), \bibinfo{pages}{102095}.
\newblock


\bibitem[Ogunkan and Ogunkan(2025)]%
        {ogunkan2025exploring}
\bibfield{author}{\bibinfo{person}{David~Victor Ogunkan} {and} \bibinfo{person}{Stella~Kehinde Ogunkan}.} \bibinfo{year}{2025}\natexlab{}.
\newblock \showarticletitle{Exploring big data applications in sustainable urban infrastructure: A review}.
\newblock \bibinfo{journal}{\emph{Urban Governance}} (\bibinfo{year}{2025}).
\newblock


\bibitem[Pearl(2009)]%
        {pearl2009causality}
\bibfield{author}{\bibinfo{person}{Judea Pearl}.} \bibinfo{year}{2009}\natexlab{}.
\newblock \bibinfo{booktitle}{\emph{Causality: Models Reasoning and Inference}}.
\newblock \bibinfo{publisher}{Cambridge University Press}.
\newblock


\bibitem[Peters et~al\mbox{.}(2016)]%
        {peters2016causal}
\bibfield{author}{\bibinfo{person}{Jonas Peters}, \bibinfo{person}{Peter B{\"u}hlmann}, {and} \bibinfo{person}{Nicolai Meinshausen}.} \bibinfo{year}{2016}\natexlab{}.
\newblock \showarticletitle{Causal inference by using invariant prediction: identification and confidence intervals}.
\newblock \bibinfo{journal}{\emph{Journal of the Royal Statistical Society Series B: Statistical Methodology}} \bibinfo{volume}{78}, \bibinfo{number}{5} (\bibinfo{year}{2016}), \bibinfo{pages}{947--1012}.
\newblock


\bibitem[Peters et~al\mbox{.}(2017)]%
        {peters2017elements}
\bibfield{author}{\bibinfo{person}{Jonas Peters}, \bibinfo{person}{Dominik Janzing}, {and} \bibinfo{person}{Bernhard Sch{\"o}lkopf}.} \bibinfo{year}{2017}\natexlab{}.
\newblock \bibinfo{booktitle}{\emph{Elements of causal inference: foundations and learning algorithms}}.
\newblock \bibinfo{publisher}{The MIT press}.
\newblock


\bibitem[Petroni et~al\mbox{.}(2019)]%
        {llmknowledgebase}
\bibfield{author}{\bibinfo{person}{Fabio Petroni}, \bibinfo{person}{Tim Rocktäschel}, \bibinfo{person}{Patrick Lewis}, \bibinfo{person}{Anton Bakhtin}, \bibinfo{person}{Yuxiang Wu}, \bibinfo{person}{AH Miller}, {and} \bibinfo{person}{Sebastian Riedel}.} \bibinfo{year}{2019}\natexlab{}.
\newblock \showarticletitle{Language Models as Knowledge Bases?}. In \bibinfo{booktitle}{\emph{Proceedings of the 2019 Conference on Empirical Methods in Natural Language Processing}}. \bibinfo{publisher}{Association for Computational Linguistics}, \bibinfo{address}{Hong Kong, China}, \bibinfo{pages}{2463--2473}.
\newblock
\href{https://doi.org/10.18653/v1/D19-1250}{doi:\nolinkurl{10.18653/v1/D19-1250}}


\bibitem[Rosseel and Loh(2022)]%
        {rosseel2022structural}
\bibfield{author}{\bibinfo{person}{Yves Rosseel} {and} \bibinfo{person}{Wen~Wei Loh}.} \bibinfo{year}{2022}\natexlab{}.
\newblock \showarticletitle{A structural after measurement approach to structural equation modeling.}
\newblock \bibinfo{journal}{\emph{Psychological Methods}} (\bibinfo{year}{2022}).
\newblock


\bibitem[Sadrizadeh et~al\mbox{.}(2022)]%
        {epa2009iaq}
\bibfield{author}{\bibinfo{person}{Sasan Sadrizadeh}, \bibinfo{person}{Runming Yao}, \bibinfo{person}{Feng Yuan}, \bibinfo{person}{Hazim Awbi}, \bibinfo{person}{William Bahnfleth}, \bibinfo{person}{Yang Bi}, \bibinfo{person}{Guangyu Cao}, \bibinfo{person}{Cristiana Croitoru}, \bibinfo{person}{Richard De~Dear}, \bibinfo{person}{Fariborz Haghighat}, {et~al\mbox{.}}} \bibinfo{year}{2022}\natexlab{}.
\newblock \showarticletitle{Indoor air quality and health in schools: A critical review for developing the roadmap for the future school environment}.
\newblock \bibinfo{journal}{\emph{Journal of Building Engineering}}  \bibinfo{volume}{57} (\bibinfo{year}{2022}), \bibinfo{pages}{104908}.
\newblock


\bibitem[Saini et~al\mbox{.}(2020)]%
        {saini2020comprehensive}
\bibfield{author}{\bibinfo{person}{Jagriti Saini}, \bibinfo{person}{Maitreyee Dutta}, {and} \bibinfo{person}{Gon{\c{c}}alo Marques}.} \bibinfo{year}{2020}\natexlab{}.
\newblock \showarticletitle{A comprehensive review on indoor air quality monitoring systems for enhanced public health}.
\newblock \bibinfo{journal}{\emph{Sustainable environment research}} \bibinfo{volume}{30}, \bibinfo{number}{1} (\bibinfo{year}{2020}), \bibinfo{pages}{6}.
\newblock


\bibitem[Seem(1987)]%
        {seem1987modeling}
\bibfield{author}{\bibinfo{person}{John~Ervin Seem}.} \bibinfo{year}{1987}\natexlab{}.
\newblock \emph{\bibinfo{title}{Modeling of heat transfer in buildings}}.
\newblock \bibinfo{thesistype}{Ph.\,D. Dissertation}. \bibinfo{school}{The University of Wisconsin-Madison}.
\newblock


\bibitem[Spirtes et~al\mbox{.}(2000)]%
        {spirtes2000causation}
\bibfield{author}{\bibinfo{person}{Peter Spirtes}, \bibinfo{person}{Clark~N Glymour}, \bibinfo{person}{Richard Scheines}, {and} \bibinfo{person}{David Heckerman}.} \bibinfo{year}{2000}\natexlab{}.
\newblock \bibinfo{booktitle}{\emph{Causation, prediction, and search}}.
\newblock \bibinfo{publisher}{MIT press}.
\newblock


\bibitem[Sun and Li(2024)]%
        {sunleveraging}
\bibfield{author}{\bibinfo{person}{Zhuofan Sun} {and} \bibinfo{person}{Qingyi Li}.} \bibinfo{year}{2024}\natexlab{}.
\newblock \showarticletitle{Leveraging LLMs for Causal Inference and Discovery}.
\newblock \bibinfo{journal}{\emph{arXiv preprint arXiv:2410.16676}} (\bibinfo{year}{2024}).
\newblock


\bibitem[Toth et~al\mbox{.}(2022)]%
        {toth2022active}
\bibfield{author}{\bibinfo{person}{Christian Toth}, \bibinfo{person}{Lars Lorch}, \bibinfo{person}{Christian Knoll}, \bibinfo{person}{Andreas Krause}, \bibinfo{person}{Franz Pernkopf}, \bibinfo{person}{Robert Peharz}, {and} \bibinfo{person}{Julius Von~K{\"u}gelgen}.} \bibinfo{year}{2022}\natexlab{}.
\newblock \showarticletitle{Active bayesian causal inference}.
\newblock \bibinfo{journal}{\emph{Advances in Neural Information Processing Systems}}  \bibinfo{volume}{35} (\bibinfo{year}{2022}), \bibinfo{pages}{16261--16275}.
\newblock


\bibitem[{U.S. Department of Energy}(2021)]%
        {DOE2021}
\bibfield{author}{\bibinfo{person}{{U.S. Department of Energy}}.} \bibinfo{year}{2021}\natexlab{}.
\newblock \bibinfo{title}{Building Energy Codes Program}.
\newblock
\urldef\tempurl%
\url{https://www.energycodes.gov/}
\showURL{%
\tempurl}
\newblock
\shownote{Accessed July 2025}.


\bibitem[{U.S. Energy Information Administration}(2018)]%
        {eia2018cbecs}
\bibfield{author}{\bibinfo{person}{{U.S. Energy Information Administration}}.} \bibinfo{year}{2018}\natexlab{}.
\newblock \bibinfo{title}{2018 Commercial Buildings Energy Consumption Survey (CBECS)}.
\newblock
\urldef\tempurl%
\url{https://www.eia.gov/consumption/commercial/}
\showURL{%
\tempurl}
\newblock
\shownote{Estimated total of 5.9 million U.S. commercial buildings surveyed}.


\bibitem[Yang et~al\mbox{.}(2014)]%
        {yang2014systematic}
\bibfield{author}{\bibinfo{person}{Zheng Yang}, \bibinfo{person}{Nan Li}, \bibinfo{person}{Burcin Becerik-Gerber}, {and} \bibinfo{person}{Michael Orosz}.} \bibinfo{year}{2014}\natexlab{}.
\newblock \showarticletitle{A systematic approach to occupancy modeling in ambient sensor-rich buildings}.
\newblock \bibinfo{journal}{\emph{Simulation}} \bibinfo{volume}{90}, \bibinfo{number}{8} (\bibinfo{year}{2014}), \bibinfo{pages}{960--977}.
\newblock


\bibitem[Ye et~al\mbox{.}(2023)]%
        {ye2023comprehensive}
\bibfield{author}{\bibinfo{person}{Junjie Ye}, \bibinfo{person}{Xuanting Chen}, \bibinfo{person}{Nuo Xu}, \bibinfo{person}{Can Zu}, \bibinfo{person}{Zekai Shao}, \bibinfo{person}{Shichun Liu}, \bibinfo{person}{Yuhan Cui}, \bibinfo{person}{Zeyang Zhou}, \bibinfo{person}{Chao Gong}, \bibinfo{person}{Yang Shen}, {et~al\mbox{.}}} \bibinfo{year}{2023}\natexlab{}.
\newblock \showarticletitle{A comprehensive capability analysis of gpt-3 and gpt-3.5 series models}.
\newblock \bibinfo{journal}{\emph{arXiv preprint arXiv:2303.10420}} (\bibinfo{year}{2023}).
\newblock


\bibitem[Yu et~al\mbox{.}(2024)]%
        {yu2024causaleval}
\bibfield{author}{\bibinfo{person}{Longxuan Yu}, \bibinfo{person}{Delin Chen}, \bibinfo{person}{Siheng Xiong}, \bibinfo{person}{Qingyang Wu}, \bibinfo{person}{Qingzhen Liu}, \bibinfo{person}{Dawei Li}, \bibinfo{person}{Zhikai Chen}, \bibinfo{person}{Xiaoze Liu}, {and} \bibinfo{person}{Liangming Pan}.} \bibinfo{year}{2024}\natexlab{}.
\newblock \showarticletitle{Causaleval: Towards better causal reasoning in language models}.
\newblock \bibinfo{journal}{\emph{arXiv preprint arXiv:2410.16676}} (\bibinfo{year}{2024}).
\newblock


\bibitem[Zapata et~al\mbox{.}(2024)]%
        {zapata2024combining}
\bibfield{author}{\bibinfo{person}{David Zapata}, \bibinfo{person}{Marcel Meyer}, {and} \bibinfo{person}{Oliver M{\"u}ller}.} \bibinfo{year}{2024}\natexlab{}.
\newblock \showarticletitle{Combining Causal Discovery and Machine Learning for Modeling Data Center Operations}.
\newblock \bibinfo{journal}{\emph{arXiv preprint arXiv:2410.09516}} (\bibinfo{year}{2024}).
\newblock


\bibitem[Ze{\v{c}}evi{\'c} et~al\mbox{.}(2023)]%
        {zevcevic2023causal}
\bibfield{author}{\bibinfo{person}{Matej Ze{\v{c}}evi{\'c}}, \bibinfo{person}{Moritz Willig}, \bibinfo{person}{Devendra~Singh Dhami}, {and} \bibinfo{person}{Kristian Kersting}.} \bibinfo{year}{2023}\natexlab{}.
\newblock \showarticletitle{Causal parrots: Large language models may talk causality but are not causal}.
\newblock \bibinfo{journal}{\emph{arXiv preprint arXiv:2308.13067}} (\bibinfo{year}{2023}).
\newblock


\bibitem[Zhang et~al\mbox{.}(2022a)]%
        {zhang2022fdd}
\bibfield{author}{\bibinfo{person}{Chaobo Zhang}, \bibinfo{person}{Yazhou Zhao}, \bibinfo{person}{Yang Zhao}, \bibinfo{person}{Tingting Li}, {and} \bibinfo{person}{Xuejun Zhang}.} \bibinfo{year}{2022}\natexlab{a}.
\newblock \showarticletitle{Causal discovery and inference-based fault detection and diagnosis method for heating, ventilation and air-conditioning systems}.
\newblock \bibinfo{journal}{\emph{Building and Environment}}  \bibinfo{volume}{212} (\bibinfo{year}{2022}), \bibinfo{pages}{108760}.
\newblock
\href{https://doi.org/10.1016/j.buildenv.2022.108760}{doi:\nolinkurl{10.1016/j.buildenv.2022.108760}}


\bibitem[Zhang et~al\mbox{.}(2022b)]%
        {zhang2022causal}
\bibfield{author}{\bibinfo{person}{Chaobo Zhang}, \bibinfo{person}{Yazhou Zhao}, \bibinfo{person}{Yang Zhao}, \bibinfo{person}{Tingting Li}, {and} \bibinfo{person}{Xuejun Zhang}.} \bibinfo{year}{2022}\natexlab{b}.
\newblock \showarticletitle{Causal discovery and inference-based fault detection and diagnosis method for heating, ventilation and air conditioning systems}.
\newblock \bibinfo{journal}{\emph{Building and Environment}}  \bibinfo{volume}{212} (\bibinfo{year}{2022}), \bibinfo{pages}{108760}.
\newblock


\bibitem[Zhang and Srinivasan(2020)]%
        {zhang2020systematic}
\bibfield{author}{\bibinfo{person}{He Zhang} {and} \bibinfo{person}{Ravi Srinivasan}.} \bibinfo{year}{2020}\natexlab{}.
\newblock \showarticletitle{A systematic review of air quality sensors, guidelines, and measurement studies for indoor air quality management}.
\newblock \bibinfo{journal}{\emph{Sustainability}} \bibinfo{volume}{12}, \bibinfo{number}{21} (\bibinfo{year}{2020}), \bibinfo{pages}{9045}.
\newblock


\bibitem[Zhang et~al\mbox{.}(2023)]%
        {zhang2023active}
\bibfield{author}{\bibinfo{person}{Jiaqi Zhang}, \bibinfo{person}{Louis Cammarata}, \bibinfo{person}{Chandler Squires}, \bibinfo{person}{Themistoklis~P Sapsis}, {and} \bibinfo{person}{Caroline Uhler}.} \bibinfo{year}{2023}\natexlab{}.
\newblock \showarticletitle{Active learning for optimal intervention design in causal models}.
\newblock \bibinfo{journal}{\emph{Nature Machine Intelligence}} \bibinfo{volume}{5}, \bibinfo{number}{10} (\bibinfo{year}{2023}), \bibinfo{pages}{1066--1075}.
\newblock


\bibitem[Zheng et~al\mbox{.}(2020)]%
        {zheng2020learning}
\bibfield{author}{\bibinfo{person}{Xun Zheng}, \bibinfo{person}{Chen Dan}, \bibinfo{person}{Bryon Aragam}, \bibinfo{person}{Pradeep Ravikumar}, {and} \bibinfo{person}{Eric Xing}.} \bibinfo{year}{2020}\natexlab{}.
\newblock \showarticletitle{Learning sparse nonparametric dags}. In \bibinfo{booktitle}{\emph{International Conference on Artificial Intelligence and Statistics}}. Pmlr, \bibinfo{pages}{3414--3425}.
\newblock


\end{thebibliography}

\appendix

\end{document}